\documentclass[10pt,twocolumn,letterpaper]{article}
\pdfoutput=1
\usepackage{cvpr}              

\usepackage[accsupp]{axessibility}  

\usepackage[pagebackref,breaklinks,colorlinks]{hyperref}

\usepackage{times}
\usepackage{epsfig}
\usepackage{graphicx}
\usepackage{amsmath}
\usepackage{amssymb}

\usepackage{booktabs}
\usepackage{commath}
\usepackage{mathtools}
\usepackage{enumerate}
\usepackage{enumitem}
\usepackage{xcolor}
\usepackage{soul}

\usepackage{caption}
\usepackage{subcaption}
\usepackage[font=small,skip=3pt]{caption}
\usepackage{physics}
\usepackage{algorithm}
\usepackage{algcompatible}
\usepackage[dvipsnames]{xcolor}
\usepackage{multirow}
\usepackage{subcaption}
\usepackage{float}

\def\paperID{10494} 
\def\confName{CVPR}
\def\confYear{2025}

\title{DiSRT-In-Bed: Diffusion-Based Sim-to-Real Transfer Framework for In-Bed Human Mesh Recovery}

\author{Jing Gao
\hspace{25pt} Ce Zheng  
 \hspace{25pt} Laszlo A. Jeni  
 \hspace{25pt} Zackory Erickson \\
 Carnegie Mellon University \\
{\tt\small \{jinggao2, cezheng, laszlojeni, zerickso\}@andrew.cmu.edu}
}

\begin{document}
\maketitle
\begin{abstract}
In-bed human mesh recovery can be crucial and enabling for several healthcare applications, including sleep pattern monitoring, rehabilitation support, and pressure ulcer prevention. However, it is difficult to collect large real-world visual datasets in this domain, in part due to privacy and expense constraints, which in turn presents significant challenges for training and deploying deep learning models. Existing in-bed human mesh estimation methods often rely heavily on real-world data, limiting their ability to generalize across different in-bed scenarios, such as varying coverings and environmental settings. To address this, we propose a Sim-to-Real Transfer Framework for in-bed human mesh recovery from overhead depth images, which leverages large-scale synthetic data alongside limited or no real-world samples. We introduce a diffusion model that bridges the gap between synthetic data and real data to support generalization in real-world in-bed pose and body inference scenarios. Extensive experiments and ablation studies validate the effectiveness of our framework, demonstrating significant improvements in robustness and adaptability across diverse healthcare scenarios.
Project page can be found at \href{https://jing-g2.github.io/DiSRT-In-Bed/}{\textcolor{red}{https://jing-g2.github.io/DiSRT-In-Bed/}}.

\end{abstract}

\section{Introduction}
\label{sec:intro}

Human mesh recovery, the process of estimating 3D human body shapes and poses from camera or sensor data, is a challenging problem with significant applications in healthcare.
In-bed human mesh recovery, in particular, plays a vital role in assessing patient well-being, monitoring mobility, and detecting health risks, such as pressure ulcers. 

However, collecting labeled real-world data in healthcare settings is costly, time-consuming, and often constrained by privacy concerns. 
Alternative sensing technologies, like pressure sensing mats, are expensive, require direct patient contact, and can lose calibration over time, limiting their reliability. Thermal sensors, while contact-free, are highly sensitive to environmental factors. In addition, both RGB cameras and thermal sensors often struggle with occlusions in bed, such as blankets. Given these limitations, depth cameras emerge as a practical solution, offering a balance of accuracy, affordability, and privacy protection, while avoiding the drawbacks of other sensor types. 

Furthermore, general human mesh recovery tasks benefit from abundant real-world data featuring individuals in standing or active poses \cite{goel2023humans,tian2023recovering}. 
Such dependence on large-scale real-world data limits the adaptability and performance of deep learning models for human mesh prediction in clinical settings, where data collection is challenging and often privacy-constrained.

\begin{figure}
    \centering
    \includegraphics[width=\linewidth]{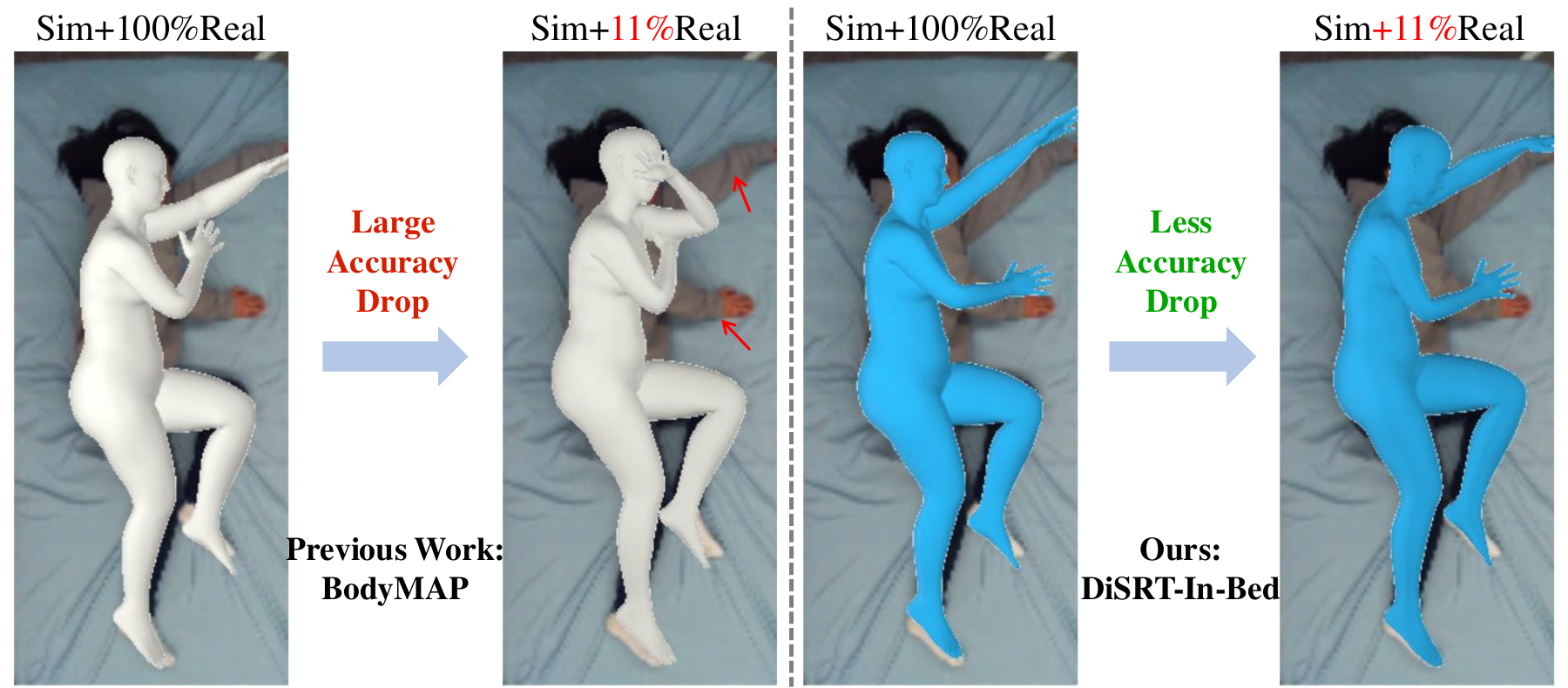}
    \caption{Impact of real-world data scarcity on in-bed human mesh recovery. BodyMAP shows significant performance degradation when trained with limited real-world data, while our method maintains robust accuracy. `Sim' indicates training with all synthetic data and `$n\%\text{Real}$' indicates training with $n\%$ of the real data from the training dataset.}
    \label{fig:teaser}
    \vspace{-15pt}
\end{figure}

To address the challenge of limited training data, utilizing synthetic data presents a promising solution. Large-scale simulated depth datasets can be efficiently generated without preserving any personally identifiable information, eliminating the need for sensitive real-world data. Building on this strategy, prior work~\cite{clever2020bodies,clever2022bodypressure,tandon2024bodymap} has demonstrated good performance in the in-bed human mesh recovery task. However, they struggle to effectively bridge the domain gap between synthetic and real-world data, leading to significant performance degradation when the proportion of real-world data in the training set is low, as shown in Fig.~\ref{fig:teaser}.
Thus, to further enhance generalization, we propose a novel diffusion-based pipeline for in-bed human mesh recovery. Diffusion models are particularly well-suited for this scenario due to their strong ability to handle uncertainties, such as noise and variations in depth images. By leveraging the diffusion framework, our method mitigates the domain gap between real and synthetic data while enhancing generalization across diverse real-world environments (e.g., different hospitals or room setups). This ensures smooth and coherent predictions across varied settings for real-world applications. We conduct extensive experiments, comparing with baselines and performing ablations, to demonstrate the method's effectiveness in real-world scenarios with occlusions and varying conditions.

Our contributions are summarized as follows:
\begin{itemize}
    \item We propose a Sim-to-Real Transfer Framework for in-bed human mesh recovery that effectively leverages synthetic data to improve performance in real-world healthcare settings with limited labeled real data.
    \item We introduce a diffusion-based architecture, enabling the diffusion process to bridge the domain gap between synthetic and real-world data and achieve strong generalization across different environment settings.
    \item We conduct extensive experiments, including comparisons with state-of-the-art methods, ablation studies, and generalization tests across different real-world settings, to assess the performance of our approach.
\end{itemize}
\section{Related Work}
\label{sec:related_work}

\subsection{3D Human Pose and Mesh Estimation}

With significant progress in 3D human pose estimation \cite{mehta2017vnect,pavllo20193d,zhu2023motionbert,zhao2024single}, researchers are seeking to go beyond just pose prediction. Human mesh recovery, which provides a more detailed 3D representation of the human body, has gained increasing interest. As a foundational human parametric model, SMPL \cite{SMPL:2015} has been widely adopted in numerous works ~\cite{kanazawaHMR18,Kolotouros2019SPIN,lin2021metro,pymaf2021,li2022cliff,zheng2023feater,wang2023refit,li2023niki,lin2023one} to recover human mesh by predicting SMPL parameters. HybrIK \cite{li2021hybrik} and its extension, HybrIK-X \cite{li2023hybrik}, introduce hybrid inverse kinematics techniques that convert 3D joints into body-part rotations through twist-and-swing decomposition. 
HMR2.0 \cite{goel2023humans} employs a straightforward yet effective transformer-based network, setting a foundation for subsequent work. TokenHMR \cite{dwivedi2024tokenhmr} introduces a tokenized approach to representing human pose and shape, effectively handling occlusions by reframing the problem as token prediction.

\subsection{Diffusion Models for Human Pose and Mesh Estimation}

Diffusion generative models \cite{ho2020denoising,song2020denoising} have shown remarkable success across diverse computer vision tasks, including image inpainting \cite{song2021scorebased,zhang2024avid}, text-to-image generation  \cite{rombach2022high,peebles2023scalable,photorealistic,zhang2023controlnet,li2025controlnetpp}, and image-to-image translation \cite{choi2021ilvr}. Leveraging their powerful capability to manage uncertainty and refine distributions, these models have been effectively applied to 3D human pose estimation and human mesh recovery tasks \cite{gong2023diffpose,cai2024disentangled,shan2023diffusion,li2023ego,stathopoulos2024scorehmr,lu2023dposer,ta2024MOPED,zheng2023diffmesh}. DiffPose \cite{gong2023diffpose} pioneers diffusion-based 3D pose prediction from 2D sequences.
Extending to human mesh recovery, HMDiff \cite{foo2023distribution} applies a distribution alignment technique to provide input-specific information within the diffusion process, simplifying mesh estimation. Similarly, ScoreHMR \cite{stathopoulos2024scorehmr} uses a diffusion model as a prior for SMPL body model parameters, guiding the denoising process with observed 2D keypoints. 

\subsection{In-Bed Human Pose and mesh Estimation}

In contrast to general human pose estimation and mesh recovery tasks, where numerous large-scale datasets are available for training, in-bed human pose estimation and mesh recovery present unique challenges. These challenges stem from the limited availability of suitable datasets, the reliance on depth images as input, and the nature of in-bed poses. Individuals are often lying in various orientations on the bed and are partially covered by blankets, leading to heavy occlusions and constrained body positions. 
As one of the prior works, Pyramid Fusion \cite{yin2022multimodal} introduces a pyramid scheme to effectively fuse four input modalities—RGB, pressure, depth, and infrared images—for human mesh estimation. However, subsequent approaches removed RGB images from the input due to privacy concerns in clinical deployments. 
PressureNet \cite{clever2020bodies} employs pressure images as input, using a multi-stage CNN-based framework to produce human mesh outputs, while BodyPressure \cite{clever2022bodypressure} focuses on depth images to infer both human mesh and pressure maps. The recent BodyMAP \cite{tandon2024bodymap} simplifies the processes used in PressureNet \cite{clever2020bodies} and BPBnet \cite{clever2022bodypressure}, predicting human mesh using depth and pressure images. In contrast, we focus on improving the generalization of the in-bed human mesh recovery from depth images by leveraging synthetic data and introducing a novel diffusion-based pipeline.

\begin{figure*}[ht]
  \centering
   \includegraphics[width=0.95\linewidth]{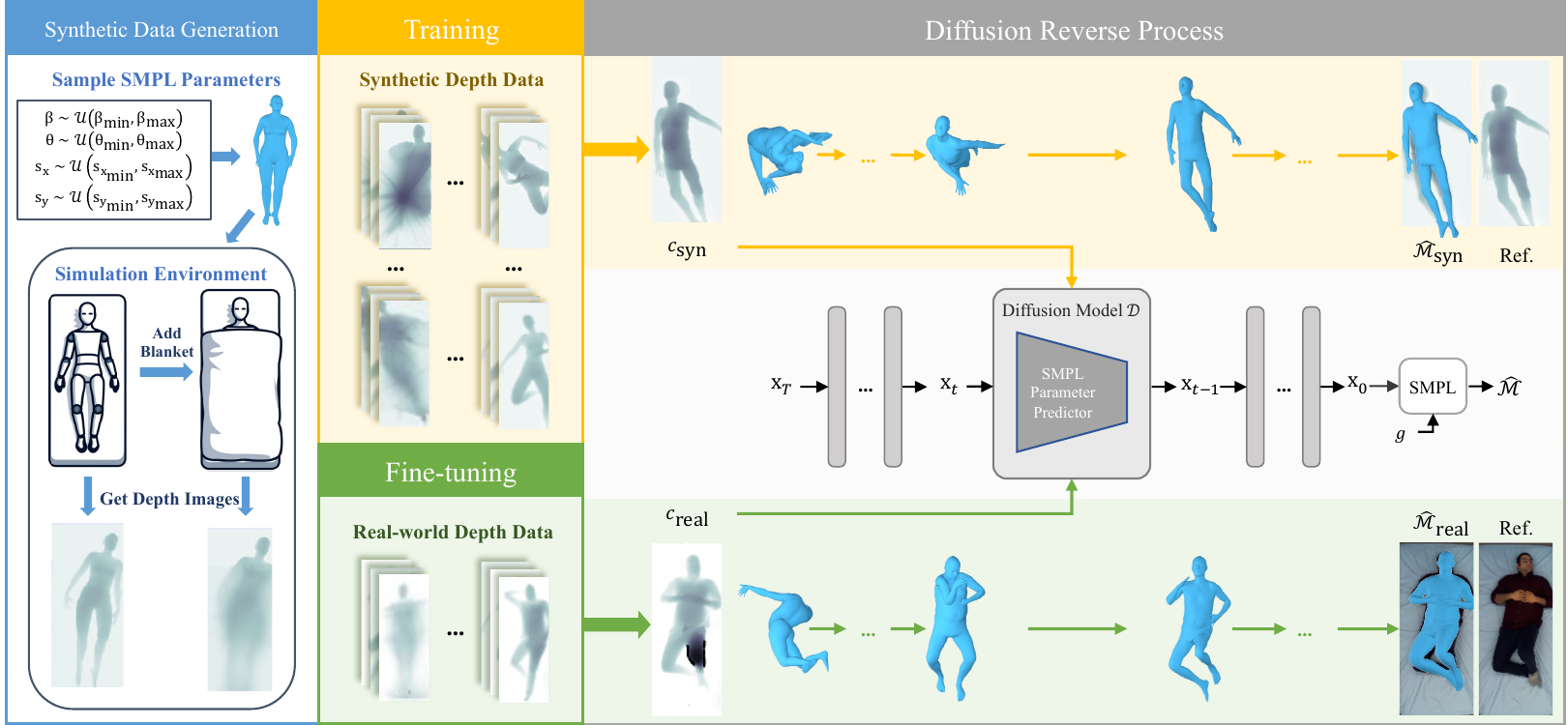}
   \caption{\textbf{Overview of the Proposed Sim-to-Real Transfer Framework.} The framework comprises three stages: In the Synthetic Data Generation stage (left), a large, diverse set of synthetic depth images is generated within a simulated environment. In the \textcolor{YellowOrange}{training} stage, the diffusion model $\mathcal{D}$ conditions on the synthetic depth image $\mathbf{c}_\text{syn}$ to denoise SMPL parameters $\mathbf{x}_t$ in the reverse process, which begins at timestep $T$ and progresses toward timestep 0, yielding the estimated human mesh $\hat{\mathcal{M}}_\text{syn}$. In the \textcolor{ForestGreen}{fine-tuning} stage, the model conditions on real depth images $\mathbf{c}_\text{real}$ to estimate the human mesh $\hat{\mathcal{M}}_\text{real}$. The symbol `$g$' in the diffusion model indicates the gender flag associated with the input. The `Ref.' in the figure denotes the corresponding synthetic depth image during training and the corresponding RGB image for visualization purposes only.}
  \vspace{-10pt}
   \label{fig:s2r_framework}
\end{figure*}

\section{Preliminaries on Diffusion Models}
\label{sec:background}

Diffusion models \cite{ho2020denoising, song2020denoising} are probabilistic generative models that learn to transform random noise into the target data distribution via a \textit{forward} and \textit{reverse} process.

In the \textbf{forward diffusion} process, a data sample $\mathbf{x}_0$ is progressively noised by adding Gaussian noise according to a fixed variance schedule $\sigma_t$ over a sequence of $T$ timesteps. This process forms a Markov chain with the transitions:
\vspace{-5pt}
\begin{equation}
q(\mathbf{x}_t \mid \mathbf{x}_{t-1}) = \mathcal{N}( \mathbf{x}_t; \sqrt{\alpha_t} \, \mathbf{x}_{t-1}, (1-\alpha_t) \mathbf{I}),
\vspace{-5pt}
\end{equation}
where $\mathbf{x}_t$ represents the noisy sample at step $t$, constant $\alpha_t = 1 - \sigma^2_t $, and $\mathcal{N}(\cdot)$ denotes the Gaussian distribution. The final sample $\mathbf{x}_t$ is approximately Gaussian noise.

The forward diffusion process defined in \cite{ho2020denoising} allows us to directly sample an arbitrary step of the noised latent $\mathbf{x}_t$ conditioned on the input $\mathbf{x}_0$ as follows:
\vspace{-5pt}
\begin{equation}
    q(\mathbf{x}_t \mid \mathbf{x}_0) = \mathcal{N}(\mathbf{x}_t; \sqrt{\bar{\alpha}_t} \, \mathbf{x}_0, (1 - \bar{\alpha}_t) \mathbf{I})
\vspace{-5pt}
\end{equation}
\begin{equation}
    \mathbf{x}_t = \sqrt{\bar{\alpha}_t} \mathbf{x}_0 + \sqrt{1 - \bar{\alpha}_t} \, \mathbf{\epsilon}, \quad \mathbf{\epsilon} \sim \mathcal{N}(0, \mathbf{I}),
    \label{eq:forward}
\vspace{-5pt}
\end{equation}
where $\alpha_t = 1 - \sigma^2_t$ and $\bar{\alpha}_t = \prod_{i=0}^{t} \alpha_i$ are fixed hyper-parameters.

In the \textbf{reverse diffusion} process, the model aims to recover the original data sample $\mathbf{x}_0$ from $\mathbf{x}_t$. A diffusion model parameterized as $\omega$ (often a neural network) is trained to approximate this reverse process defined as:
\vspace{-5pt}
\begin{equation}
    p_\omega(\mathbf{x}_{t-1} \mid \mathbf{x}_t) = \mathcal{N}(\mathbf{x}_{t-1}; \mu_\omega(\mathbf{x}_t, t), \sigma_t^2 \mathbf{I}),
    \label{eq:reverse}
\vspace{-5pt}
\end{equation}

Although specific formulations for the estimated mean $\mu_\omega(\mathbf{x}_t, t) $ vary \cite{ho2020denoising, song2020denoising, nichol2021improved}, each reverse denoising step can be expressed as a function $f$ of $\mathbf{x}_t$ and the diffusion model $\omega$ to yield $\mathbf{x}_{t-1}$ as follows:
\vspace{-5pt}
\begin{equation}
\mathbf{x}_{t-1} = f(\mathbf{x}_t, \omega).
\vspace{-5pt}
\end{equation}

During inference, Gaussian noise $\mathbf{x}_t$ is sampled, and the model iteratively denoises it to generate the target sample $\mathbf{x}_0$ using the trained diffusion model $\omega$.

\section{Methodology}
\label{sec:method}

Our proposed framework addresses the challenge of developing reliable and generalizable in-bed human mesh recovery models in scenarios with limited or no real-world data. By leveraging a large volume of synthetic data generated through simulation, combined with a small amount of real-world data, our framework effectively reduces the reliance on costly and privacy-sensitive real-world data collection. The framework comprises three key stages: synthetic data generation (Sec.~\ref{subsec:data_generation}), model design (Sec.~\ref{subsec:diffusion}), and pipeline training and fine-tuning (Sec.~\ref{subsec:training}). The overview pipeline is shown in Fig. ~\ref{fig:s2r_framework}.

Throughout our approach, we utilize the SMPL~\cite{SMPL:2015} model to represent 3D human bodies. The SMPL model is a parametric human body model that represents a human figure as a mesh of vertices, controlled by a set of pose and shape parameters. Specifically, given the joint angles $\boldsymbol{\theta} \in \mathbb{R}^{23 \times 3}$ and shape parameters $\boldsymbol{\beta} \in \mathbb{R}^{10}$, the SMPL model can output a 3D human mesh 
$\mathbf{V} \in \mathbb{R}^{6890 \times 3}$ consisting of $6,890$ vertices. The SMPL parameters can be defined as $\mathbf{\mathbf{x}} = \begin{bmatrix} \boldsymbol{\beta} & \boldsymbol{\theta} & \mathbf{s} & \mathbf{u} & \mathbf{v} \end{bmatrix}^\top \in \mathbb{R}^{88}$, where $\mathbf{s}\in \mathbb{R}^{3}$ is the global translation, and $\mathbf{u} \in \mathbb{R}^3$ with $\mathbf{v} \in \mathbb{R}^3$ are used to represent the global rotation. 

\subsection{Synthetic Data Generation}
\label{subsec:data_generation}

Obtaining labeled data for in-bed scenarios across diverse healthcare environments presents a significant challenge, limiting the deployment of deep learning models in this domain. In contrast, simulation offers a low-cost and efficient solution for generating abundant, high-quality depth data along with ground truth annotation for human mesh in resting positions. By incorporating prior information such as bed dimensions and camera-to-bed distance, we can construct simulated environments that closely replicate the real-world settings.

Following BodyPressure ~\cite{clever2022bodypressure}, which introduces a physics-based simulation pipeline to generate synthetic in-bed human depth and pressure images, we adopt this approach to create a diverse and realistic dataset. The pipeline simulates human bodies at rest on a soft mattress, producing depth data from a fixed camera position relative to SMPL-based body configurations on a bed. By sampling human shape $\boldsymbol{\beta}$, joint angles $\boldsymbol{\theta}$, and global translation $(s_x, s_y)$ from uniform distributions, we generate a variety of data. Additionally, simulated depth images with blankets are created by draping various types of blankets over parts of the body. This dataset further includes diverse human shapes, poses, and bed scene complexities.

While synthetic data generation can enhance dataset diversity and increase the number of training samples, it also introduces an inherent domain gap between synthetic and real-world data. As a result, models may perform well in synthetic settings but struggle in real-world applications, which undermines their practical utility. Therefore, in the following sections, we focus on bridging this simulation-to-reality gap within the framework for in-bed scenes.

\subsection{Diffusion-Based In-Bed Mesh Recovery}
\label{subsec:diffusion}

Recovering in-bed human mesh from depth images is not a straightforward one-to-one mapping problem, as prior works ~\cite{clever2020bodies,clever2022bodypressure,tandon2024bodymap} state.
Depth images of in-bed scenarios can vary significantly based on external conditions, such as the presence or absence of a blanket, while the underlying body pose and shape remain the same. This variability introduces ambiguity in the mapping from depth images to human mesh. 
Additionally, multiple plausible human mesh configurations can correspond to the same depth image due to inherent ambiguities.
To address this, we reformulate the in-bed human mesh recovery task as a conditional generative problem. Inspired by recent advancements in diffusion models for image generation~\cite{ho2020denoising,song2020denoising}, we design a diffusion model to learn the distribution of plausible SMPL body configurations, denoted as $p_\text{SMPL}$, conditioned on depth images during training and fine-tuning. 

\subsubsection{Diffusion Process}
In contrast to the diffusion process used in image generation, which operates directly on images, we conduct forward noise-adding and reverse denoising processes on the SMPL body parameters $\mathbf{x}$ for in-bed human mesh recovery.

In the \textit{forward} process, we follow the Eq.~\ref{eq:forward} to obtain noisy versions $\mathbf{x}_t$ of the initial SMPL body parameters $\mathbf{x}_0$ over $t$ timesteps.

In the \textit{reverse} process, we incorporate depth images $\mathbf{c}$ as a conditional input to the diffusion model, modifying Eq.~\ref{eq:reverse} as follows:
\vspace{-5pt}
\begin{equation}
p_{\mathcal{D}}(\mathbf{x}_{t-1} \mid \mathbf{x}_t, \mathbf{c}) = \mathcal{N}(\mathbf{x}_{t-1}; \mu_{\mathcal{D}}(\mathbf{x}_t, t,  \mathbf{c}), \sigma_t^2 \mathbf{I}),
\label{eq:reverse_cond}
\vspace{-5pt}
\end{equation}
where $\mathcal{D}$ represents our diffusion model designed for the in-bed human mesh recovery task.

Given a training sample $\mathbf{x}_0$, we train the diffusion model $\mathcal{D}$ to learn the denoising transition $p_\mathcal{D}(\mathbf{x}_{t-1} \mid \mathbf{x}_{t}, \mathbf{c})$, ensuring it closely approximates the corresponding forward process $q(\mathbf{x}_t \mid \mathbf{x}_{t-1}, \mathbf{x}_0)$. In image generation tasks, this process typically involves having the diffusion model approximate the noise term $\epsilon$ that produces $\mathbf{x}_t$ from $\mathbf{x}_0$ in the forward process. 

However, in our case, if we assume SMPL body parameters $\mathbf{x}_t$ follow a standard Gaussian distribution, the diffusion model struggles to produce reasonable SMPL parameters, as early denoising iterations may yield unfeasible human meshes. To address this, we diffuse toward the initial sample $\mathbf{x}_0$ to ensure consistency throughout the denoising process. For any timestep $t$, the estimated initial SMPL parameters $\mathbf{z}_t$ can be represented as:
\vspace{-5pt}
\begin{equation}
\mathbf{z}_t = \mathcal{D}(\mathbf{x}_t, t, \mathbf{c})
\vspace{-5pt}
\end{equation}
The objective of training and fine-tuning the diffusion model $\mathcal{D}$ for in-bed human mesh recovery is to minimize
\vspace{-5pt}
\begin{equation}
\mathbb{E}_{\mathbf{x}_0 \sim p_\text{SMPL}} \mathbb{E}_{t \sim \mathcal{U} \{0, T\}, \mathbf{x}_t \sim q(\cdot | \mathbf{x}_0)}  \| \mathbf{z}_t - \mathbf{x}_0 \|,
\vspace{-2pt}
\end{equation}
where $\mathcal{U}(\cdot)$ denotes sample from a uniform distribution.

In the inference, the learned mean $\mu_\mathcal{D}$ in the Eq.~\ref{eq:reverse_cond} can be formulated as:
\vspace{-5pt}
\begin{equation}
\mu_\mathcal{D} = \frac{\sqrt{\bar{\alpha}_{t-1}}(1 - \alpha_t)}{1-\bar{\alpha}_t} \mathbf{z}_t + 
\frac{\sqrt{\alpha_t}(1-\bar{\alpha}_{t-1})}{1-\bar{\alpha}_t}\mathbf{x}_t
,
\vspace{-5pt}
\end{equation}
where $\alpha_t$ and $\bar\alpha_{t-1}$ are derived from hyper-parameters. Then, we sample from the transition distribution in each denoising step to compute $\mathbf{x}_{t-1}$ as follows:
\vspace{-5pt}
\begin{equation}
    \mathbf{x}_{t-1} = \mu_\mathcal{D}(\mathbf{x}_t, t, \mathbf{c}) + \sigma^2 \epsilon
\vspace{-5pt}
\end{equation}
Following this reverse process, we iteratively denoise the SMPL latent from noise $\mathbf{x}_T$ at timestep $T$ down to the target SMPL latent $\mathbf{x}_0$ at timestep 0.
\begin{figure}[ht]
  \centering
   \includegraphics[width=\linewidth]{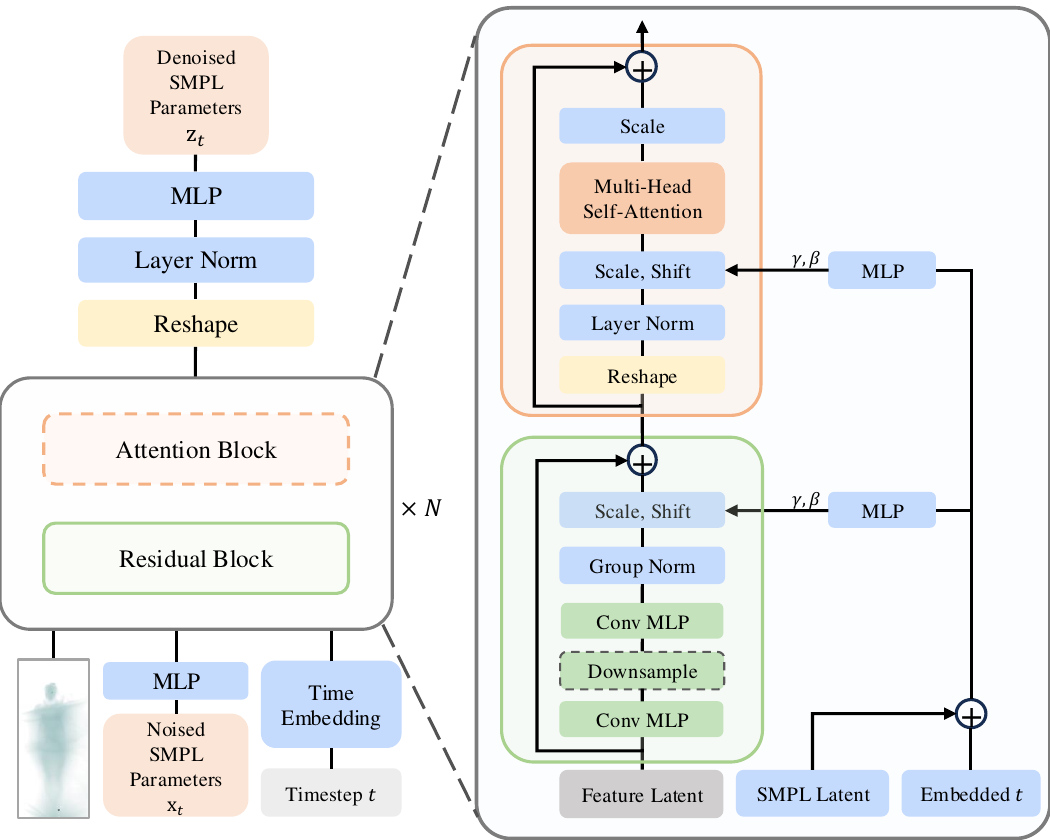}
   \caption{\textbf{Diffusion Model Architecture.} Dashed lines around specific layers indicate optional layers that may be omitted in certain blocks of the model implementation.
}
   \label{fig:architecure}
  \vspace{-10pt}
\end{figure}

\subsubsection{Model Architecture}
\label{subsubsec:arch}

We introduce a network that takes noised SMPL parameters $\mathbf{x}_t$, depth images $\mathbf{c}$, and the timestep $t$ as inputs and outputs the denoised SMPL parameters  $\mathbf{z}_t$  
as illustrated in Fig.~\ref{fig:architecure}. 
The reverse diffusion process leverages the depth feature latent to infer denoised SMPL parameters.
The noisy SMPL parameters $\mathbf{x}_t$ are processed through an MLP encoder to obtain the SMPL parameter latent, and a uniform sampler generates the time embedding for the timestep $t$. These inputs—SMPL parameter latent, depth images, and time embedding—are then processed through residual and attention blocks.

Following the design of the diffusion U-Net model ~\cite{rombach2022high}, we incorporate residual and attention blocks to process image inputs and replace batch and layer normalization with adaptive normalization layers initialized to zero ~\cite{peebles2023scalable} across the network. 
Specifically, the SMPL latent is aligned to have the same dimension as the time embedding, allowing them to be combined and fed into each adaptive normalization layer to produce scale and shift parameters, $\gamma$ and $\beta$, through a linear MLP. This approach enables dynamic adjustments of normalization parameters, enhancing the network's ability to handle conditional information effectively in diffusion models.

Within the network, multiple down-sampling residual blocks follow the initial convolutional layer, succeeded by two additional residual blocks. Attention blocks are attached separately before the last few residual blocks.
In the final block, current noised SMPL latent and embedded $t$ are omitted in subsequent layers, allowing the regressor to map the single output latents from all intermediate blocks to the target SMPL parameters.

Following the SMPL parameter predictor module, the SMPL model ~\cite{SMPL:2015} returns the human mesh $\hat{\mathcal{M}}$ given the estimated SMPL parameters $\mathbf{z}_t$. Additionally, a set of gender flags $\mathbf{g} \in \mathbb{R}^2$ controls the gender of the generated human mesh. In this work, we only have two gender flags, where $[0, 1]$ represents the female SMPL model and $[1, 0]$ the male model. The 3D Cartesian joint positions $\hat{\mathbf{J}} \in \mathbb{R}^{24 \times 3}$ can be extracted from the human mesh vertices $\hat{\mathbf{V}} \in \mathbb{R}^{6890 \times 3}$ of each human mesh $\hat{\mathcal{M}}$.

\subsection{Training Strategy}
\label{subsec:training}
As described above, the output of our diffusion model $\mathcal{D}$ consists of the estimated in-bed SMPL body parameters $\mathbf{z}_t = \begin{bmatrix} \hat{\boldsymbol{\beta}} & \hat{\boldsymbol{\theta}} & \hat{\mathbf{s}} & \hat{\mathbf{u}} & \hat{\mathbf{v}} \end{bmatrix}^\top$, which includes the predicted body shape parameters $\hat{\boldsymbol{\beta}}$, joint angles $\hat{\boldsymbol{\theta}}$, global translation $\hat{\mathbf{s}}$, and global rotation parameters 
$\hat{\mathbf{u}} = \{u_x, u_y, u_z\}$ and $\hat{\mathbf{v}} = \{v_x, v_y, v_z\}$. Each rotation component $\phi_i$ for $i \in \{x, y, z\}$ can be calculated as $\phi_i = \text{atan2}(u_i, v_i)$.

\textbf{Synthetic training stage:} Previous works ~\cite{clever2022bodypressure, tandon2024bodymap} rely on joint training with both synthetic and real-world data, assuming ample real-world data is available and overlooking the effects of the large synthetic-to-real data imbalance. In contrast, our framework decouples training on synthetic and real-world data to address the limited availability of real-world data and the high ratio of synthetic to real samples. During the synthetic data training phase, our goal is to establish a strong prior based on the diverse range of human resting postures available in the synthetic dataset, enabling the model to produce a reasonable coarse in-bed human pose without requiring real-world data. Our proposed diffusion-based network is trained for a reasonable number of steps on synthetic depth data alone, using a fixed learning rate to prevent convergence to local optima.

\textbf{Fine-tuning stage:} Given the variable quantity of real-world data, we employ a linearly adjusted learning rate scheduler that automatically adapts based on the amount of available depth data. This adaptive learning rate strategy facilitates rapid convergence and enhances generalization to real-world scenarios during fine-tuning, as demonstrated by our ablation study in Sec ~\ref{sec:ablation}.

\textbf{Loss:} The total loss used to train and fine-tune the diffusion model contains two components: SMPL parameter loss and vertex position loss. An expansion of each term in the loss function can be found in the supplementary material.

\begin{equation}
    \mathcal{L_{\text{total}}} = \mathcal{L}_{\text{SMPL}} + \lambda_{\text{v2v}} \mathcal{L}_{\text{v2v}},
\end{equation}
where $\lambda_{\text{v2v}}$ is a tunable hyper-parameter. 
\begin{table*}[ht]
\vspace{-5pt}
\centering
\resizebox{\textwidth}{!}{%
\begin{tabular}{lcccccccccccc}
    \toprule
\multirow{1}{*}{Data Split} & \multicolumn{2}{c}{Sim} & \multicolumn{2}{c}{Sim+11\%Real} & \multicolumn{2}{c}{Sim+24\%Real} & \multicolumn{2}{c}{Sim+37\%Real} & \multicolumn{2}{c}{Sim+49\%Real} & \multicolumn{2}{c}{Sim+100\%Real} \\
\multirow{1}{*}{Real-Sim Ratio} & \multicolumn{2}{c}{0:97495} & \multicolumn{2}{c}{1:80} & \multicolumn{2}{c}{1:38} & \multicolumn{2}{c}{1:25} & \multicolumn{2}{c}{1:18} & \multicolumn{2}{c}{1:9} \\
\cmidrule(lr){2-3}\cmidrule(lr){4-5}\cmidrule(lr){6-7}\cmidrule(lr){8-9}\cmidrule(lr){10-11}\cmidrule(lr){12-13}
   Method       & MPJPE       & PVE       & MPJPE          & PVE         & MPJPE          & PVE         & MPJPE          & PVE         & MPJPE          & PVE         & MPJPE          & PVE         \\
    \midrule
HMR2.0~\cite{goel2023humans}           & 157.23     & 182.27     & 87.09        & 104.65        & 82.85        & 96.54         &    145.35     &    140.27           &   90.90   &   108.90   & 76.94         & 90.39        \\
BodyPressure~\cite{clever2022bodypressure}      & \textbf{103.47}    & \textbf{115.31}     & 89.87        & 104.67        & 90.46        & 105.97        & 81.29         & 99.05        & 78.25         & 94.65        & 72.93         & 86.78        \\
BodyMAP~\cite{tandon2024bodymap}           & 330.52     & 365.43     & 85.79        & 90.14         & 76.07        & 89.80         & 69.51         & 83.01        & 61.77         & 74.96        & 57.06         & 69.95        \\
\textbf{DiSRT-In-Bed(Ours)}     & 109.73     & 121.59     & \textbf{74.37}        & \textbf{78.03}         & \textbf{67.14}        & \textbf{73.18}         & \textbf{58.04}         & \textbf{66.66}        & \textbf{55.94}         & \textbf{64.14}        & \textbf{50.81}         & \textbf{61.18}        \\ 
    \bottomrule
\end{tabular}}
    \caption{\textbf{Comparison to Baselines across Different Data Splits.} In the `Data Split' row, `Sim' indicates training with all synthetic data, while `$n\%\text{Real}$' indicates training with $n\%$ of the real data from the SLP training dataset. In the `Real-Sim Ratio' row, the number represents the approximate ratio of depth images between synthetic and real datasets. All values in the table are in millimeters (mm).}
  \label{tab:s2r_cmp}
\end{table*}
\begin{table*}[ht]
\small
\centering
\begin{tabular}{lcccccccccc}
    \toprule
\multirow{2}{*}{} & \multicolumn{2}{c}{Uncover} & \multicolumn{2}{c}{Cover 1} & \multicolumn{2}{c}{Cover 2} & \multicolumn{4}{c}{3D Shape Error(cm)$\downarrow$ } \\
\cmidrule(lr){2-3}\cmidrule(lr){4-5}\cmidrule(lr){6-7}\cmidrule(lr){8-11}
   Method     & MPJPE         & PVE         & MPJPE         & PVE         & MPJPE         & PVE         & height   & Chest  & Waist  & Hips  \\
    \midrule
HMR2.0~\cite{goel2023humans}           & 69.67          & 81.66          & 79.86          & 93.74          & 81.29          & 95.76          & 5.41          & 8.30          & 11.24         & 7.66          \\
BodyPressure~\cite{clever2022bodypressure}      & 67.06          & 79.92          & 76.39          & 90.78          & 75.36          & 89.65          & 3.96          & 3.89          & 4.84          & \textbf{3.37}          \\
BodyMAP~\cite{tandon2024bodymap}           & 51.26          & 62.34          & 60.35          & 73.97          & 59.55          & 73.54          & 3.43          & \textbf{3.17}          & \textbf{4.24}          & 3.53          \\
\textbf{DiSRT-In-Bed(Ours)}     & \textbf{46.01} & \textbf{55.07} & \textbf{53.78} & \textbf{64.80} & \textbf{52.65} & \textbf{63.68} & \textbf{3.25} & 5.17 & 7.16 & 5.20 \\
    \bottomrule
\end{tabular}
    \caption{\textbf{Comparison to Baselines across Different Covering Situations.} `Uncover' refers to testing depth images without coverings, `Cover 1' denotes images with a thin blanket, and `Cover 2' denotes images with a thick blanket. All MPJPE and PVE values are in millimeters (mm), while 3D Shape Errors are in centimeters (cm).}
  \label{tab:cover_cmp}
  \vspace{-5pt}
  \vspace{-5pt}
\end{table*}
\begin{figure*}[htp]
\vspace{-16pt}
  \centering
  \includegraphics[width=0.92\linewidth]{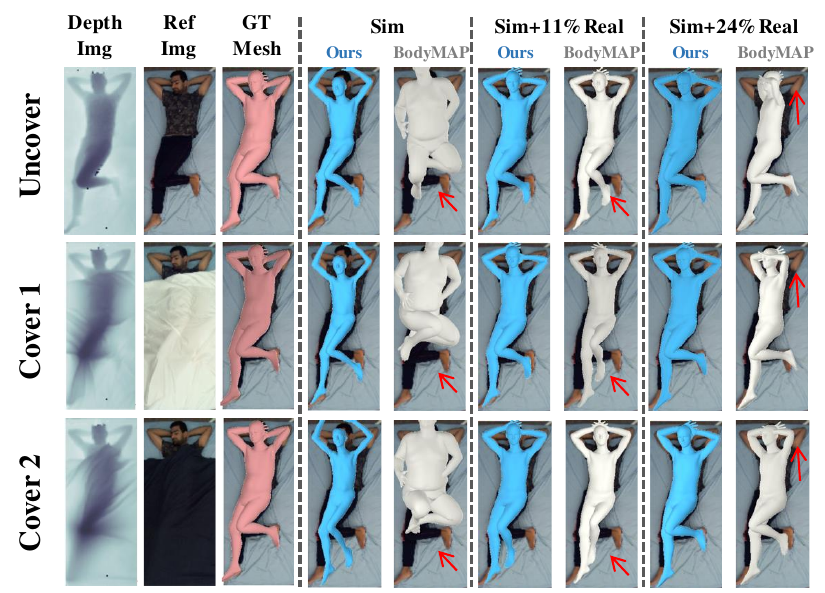}
  \caption{\textbf{Visualization of Human Mesh Estimated from limited Real-World Data in Home Settings.} The left three columns show the input depth images, RGB reference images, and ground truth mesh respectively. 
  `Sim' denotes using all simulation data in the training stage, `n\%Real' denotes the `n' ratio of real data used in the fine-tuning in our method and jointly training in the baseline. 
  `Uncover' refers to no blanket in the bed, `Cover 1' indicates the participant is covered with a thin blanket, and `Cover 2' means the participant is covered with a thick blanket. The red arrows in the figures point out the mismatch between mesh prediction and the reference images.}
  \label{fig:sim-to-real}
  \vspace{-10pt}
\end{figure*}
\section{Experiments}
\label{sec:experiments}
\subsection{Datasets and Metrics}
\label{sec:data_metrics}
\textbf{Simultaneously-collected multimodal Lying Pose (SLP)} ~\cite{slpdata} provides a comprehensive collection of in-bed resting poses across two settings. 
In the home setting, data were collected from 102 human participants, with each pose captured under three occlusion conditions: thin sheet, thicker blanket, and no covering. Clever et al.~\cite{clever2022bodypressure} provide SMPL ground truth labels for the real-world SLP home setting data.
For training, data from the first 80 participants (1-80, excluding participant 7 due to calibration errors, totaling 10,665 real samples) in the SLP are used either partially or fully during fine-tuning to represent different synthetic-to-real data ratios. For evaluation, data from the remaining 22 participants (81-102, with 2,970 real samples) are used to assess all methods.
Additionally, the SLP dataset includes data from a hospital setting, comprising 7 participants without SMPL ground truth labels. Thus, we evaluate our method on the hospital setting data through qualitative visualizations.

\textbf{BodyPressureSD}~\cite{clever2022bodypressure} is generated using the physical simulation mentioned in Sec.~\ref{subsec:data_generation} and serves as a benchmark for sim-to-real tasks. The dataset consists of 97,495 samples, corresponding to the three covering conditions in the SLP dataset, and significantly increases the diversity of human resting poses and body shapes. All synthetic data from BodyPressureSD are used during the training stage of the sim-to-real framework to establish a strong prior on human pose and shape before fine-tuning.

\textbf{Metrics.}
For pose and shape accuracy evaluation, we report the 3D mean-per-joint position error (MPJPE) and 3D per-vertex error (PVE). For each sample, MPJPE is calculated as the Mean Euclidean Distance between the inferred and ground truth positions of 24 3D joints, while PVE measures the mean Euclidean distance across the 6,890 3D vertex positions of the SMPL model.

\subsection{Implementation Details}
The diffusion model architecture consists of 6 downsampling layers within residual blocks, matching the downsampling depth of ResNet18~\cite{he2016deep}, and includes three attention blocks positioned before the final three residual blocks. The number of diffusion timesteps is set to 100 during training. For data augmentation, we shuffle the input depth images with random rotations, random erasures, and random noise additions during both training and fine-tuning, using a batch size of 32. All models are optimized with the AdamW~\cite{loshchilov2017decoupled} optimizer, using an initial learning rate of $1 \times 10^{-4}$ and a weight decay of $5 \times 10^{-4}$, trained on a single NVIDIA GeForce RTX 4090 GPU. We set $\lambda_{\text{v2v}} = 1$ in the diffusion loss.
For testing, we employ a DDIM sampling strategy with 5 timesteps to accelerate inference.

\subsection{Comparison to State-of-the-Art Methods}

We conduct several experiments to demonstrate the effectiveness of our method for in-bed human mesh recovery. We choose the current SOTA model HMR2.0~\cite{goel2023humans}, which is designed for single-person mesh recovery from general RGB images, as a baseline. 
Since our task involves single-channel depth images as input, we modify HMR2.0 by repeating the depth image across channels to match the RGB input format, making it compatible with our scenario. 
We also compare our method with BodyMAP~\cite{tandon2024bodymap} and BodyPressure~\cite{clever2022bodypressure}, designed specifically for the in-bed scenario. Tab.~\ref{tab:s2r_cmp} presents quantitative comparisons between our method and baselines across different data splits.

\noindent \textbf{\(\blacktriangle\) Generalization with limited real data:} With high real-to-simulation ratios, our method consistently outperforms previous methods in MPJPE and PVE metrics. Notably, our approach reduces these errors by over 10\% under extreme real-simulation data ratios, such as 1:80 and 1:38. When trained solely on simulation data, our model slightly underperforms BodyPressure, as BodyPressure uses a separate network for human shape parameter prediction. However, BodyPressure's approach introduces a strong bias, leading to degraded performance when limited real-world data is available. Our method demonstrates strong generalization when limited real-world data is available.

\noindent \textbf{\(\blacktriangle\) Robustness across occlusion:} As shown in Tab.~\ref{tab:cover_cmp}, we compare models trained on all simulation and real training data across various covering conditions. Our method achieves improved mesh recovery accuracy in all cases compared to prior literature and baselines. Moreover, the stability of our results (less performance drop) across varying degrees of occlusion underscores the robustness of our method, making it more reliable in real-world healthcare settings where patients are often partially covered.

\noindent \textbf{\(\blacktriangle\) Visualization:} Fig.~\ref{fig:sim-to-real} presents the visual comparisons between our method and the SOTA in-bed mesh recovery method BodyMAP~\cite{tandon2024bodymap} under different covering conditions and real-data availability.
BodyMAP struggles to estimate accurate human meshes when trained on simulation data alone, highlighting its limitations in addressing simulation-to-real domain gaps. In contrast, our method can capture meaningful pose information. Further, as shown in cases with 11\% and 24\% real data, our method’s predictions align more closely with input images across all covering scenarios. With 24\% real data, our model remains largely unaffected by coverings, consistently aligning well with the reference image and ground truth.



\subsection{Generalization to Different Real-World Settings}
\begin{figure*}[ht]
  \vspace{-5pt}
  \centering
  \subcaptionbox{Ablation on Synthetic Data\label{fig:ablation_mpjpe_syn}}{%
    \includegraphics[width=0.32\linewidth]{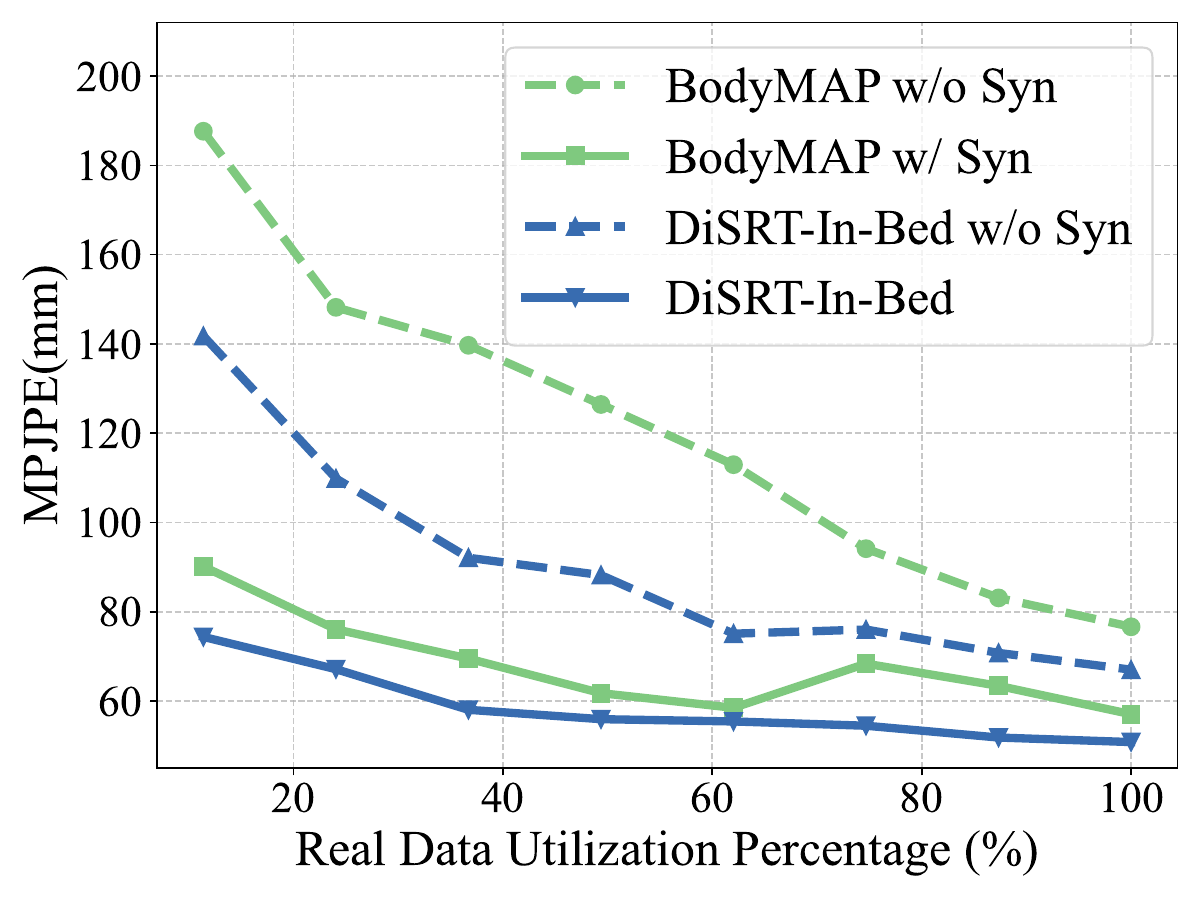}
  }
  \subcaptionbox{Ablation on Training Strategies\label{fig:ablation_mpjpe_s2r}}{%
    \includegraphics[width=0.32\linewidth]{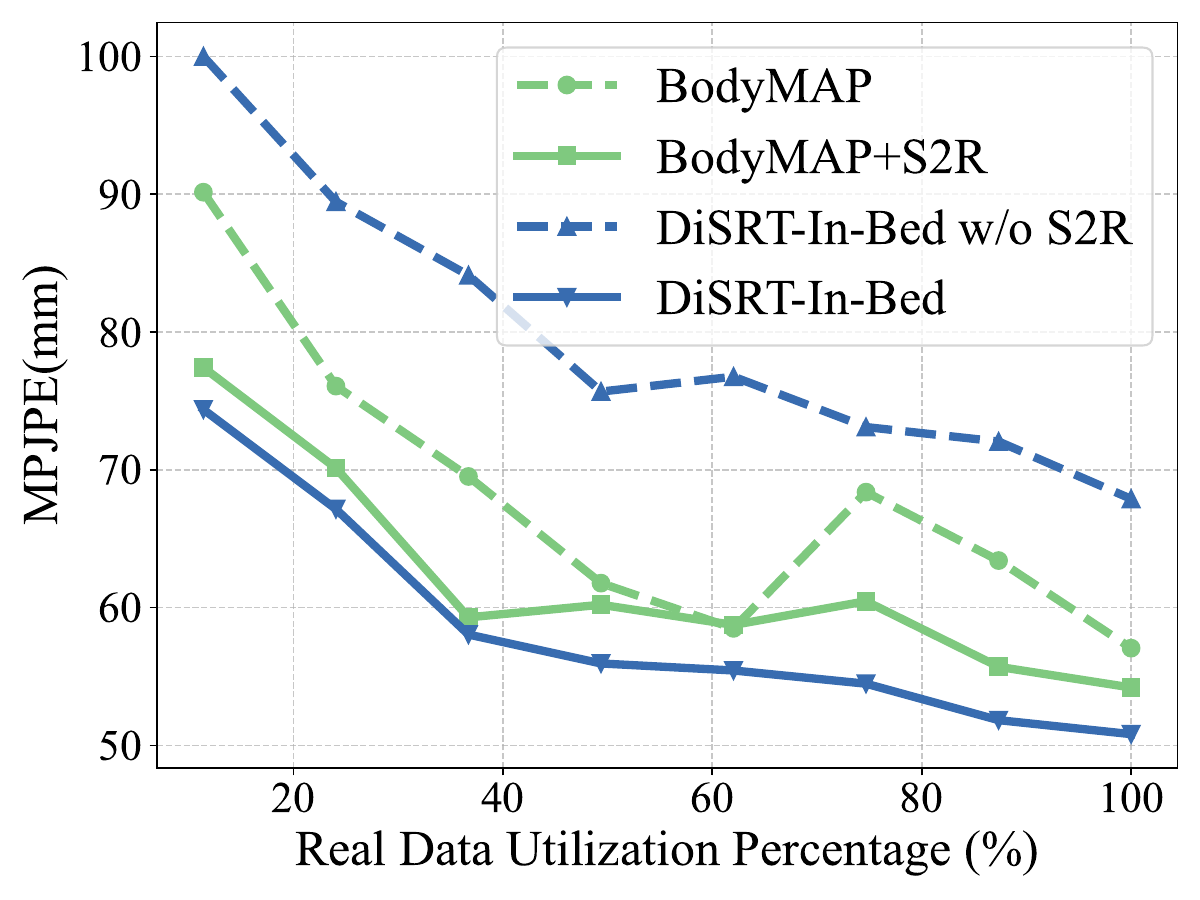}
  }
  \subcaptionbox{Ablation on Different Architectures\label{fig:ablation_mpjpe_arch}}{%
    \includegraphics[width=0.32\linewidth]{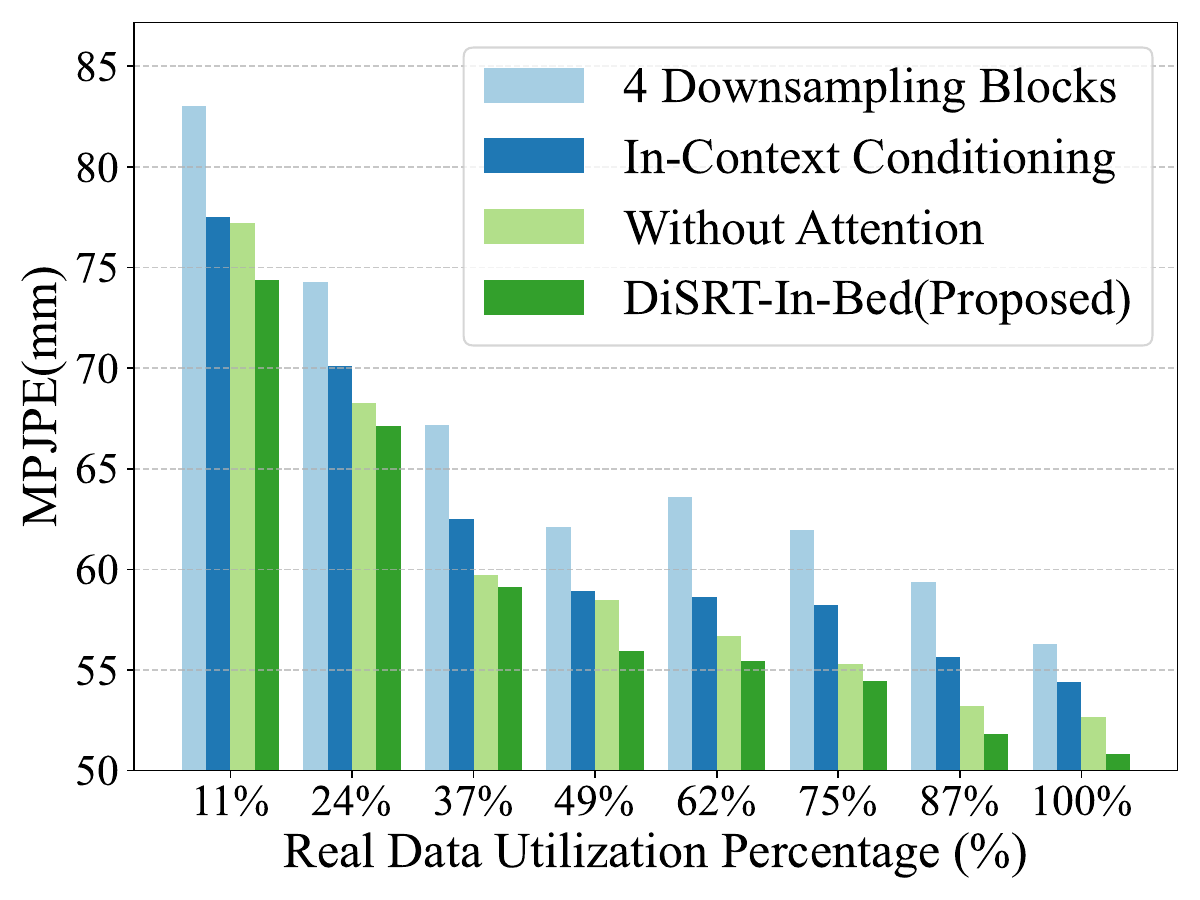}
  }
  \caption{\textbf{Ablation Study on Diffusion-Based Sim-to-Real Transfer Framework.}}
  \label{fig:ablation}
  \vspace{-5pt}
  \vspace{-5pt}
\end{figure*}
\begin{figure}[t]
\vspace{-5pt}
  \centering
  \includegraphics[width=0.92\linewidth]{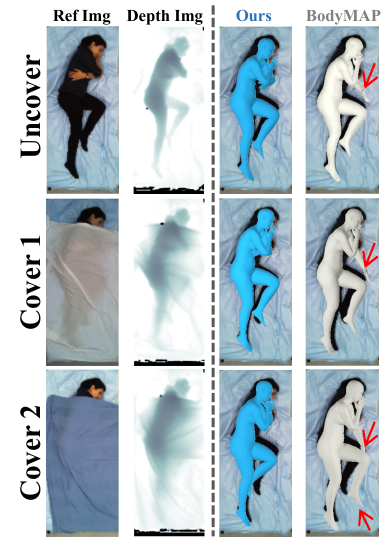}
  \caption{\textbf{Visualization of Human Mesh estimated from the Real-World Data in Hospital Settings.} Ground truth mesh is not available for this scenario. The red arrows in the figures point out the mismatch between mesh prediction and the reference images.}
  \label{fig:vis_hospital}
  \vspace{-5pt}
  \vspace{-5pt}
\end{figure}
%
To evaluate our method’s generalization ability across different environment settings, we compare our `Sim+100\%Real' model with BodyMAP\cite{tandon2024bodymap} on unlabeled hospital-setting depth images\ref{sec:data_metrics}. Since ground truth meshes are not available for this setting, we only provide qualitative comparisons. As shown in Fig.~\ref{fig:vis_hospital}, our method recovers more accurate human meshes compared to BodyMAP~\cite{tandon2024bodymap}. Notably, for the challenging self-hugging pose, our method accurately captures the crossed arms, whereas BodyMAP~\cite{tandon2024bodymap} fails to replicate this detail. Additional generalization experimental results are available in the supplementary material.

\subsection{Ablation Study}
\label{sec:ablation}

\subsubsection{Effectiveness of Sim-to-Real Transfer Framework}
\label{subsec:ablation_s2r}
To highlight the benefits of synthetic data, we plot MPJPE and PVE for our method and baselines~\cite{clever2022bodypressure,tandon2024bodymap} trained with and without synthetic data, under various real-data availability settings, in Fig.~\ref{fig:ablation_mpjpe_syn}. Row-wise comparisons reveal that adding synthetic data significantly enhances performance across all real-world data utilization percentages, especially for the baselines. Our diffusion-based framework further enhances predictions by 10-35\% in MPJPE, even when only trained on a small portion of real data.

Additionally, Fig.~\ref{fig:ablation_mpjpe_s2r} illustrates the impact of applying our Sim-to-Real (S2R) training strategy as detailed in ~\ref{subsec:training}. Once BodyMAP incorporates our S2R training strategy, it achieves a substantial improvement over its original training scheme. Moreover, our method further outperforms BodyMAP+S2R, underscoring the combined benefit of the proposed framework and model architecture in handling generalization challenges for the in-bed scenario.

\subsubsection{Effectiveness of Diffusion Model Architecture}
\label{subsec:ablation_arch}
We compare our diffusion model architecture (DiSRT-In-Bed) with three design choices as illustrated in Fig.~\ref{fig:ablation_mpjpe_arch}.

\noindent \textbf{Design Choice 1} — Number of Downsampling Blocks — 6 v.s. 4: Our final model configuration includes 6 convolutional downsampling layers, similar to the ResNet-18 architecture ~\cite{he2016deep} used in BodyMAP ~\cite{tandon2024bodymap}. We observe a decrease in MPJPE when using 6 downsampling blocks rather than only 4.

\noindent \textbf{Design Choice 2} — Conditioning Technique — Conditioning with adaptive normalization v.s. in-context conditioning: We evaluate our conditioning approach (Section \ref{subsubsec:arch}) against the commonly used in-context conditioning techniques~\cite{zhang2023controlnet,li2025controlnetpp}, which concatenates depth features and SMPL latent representations of equal size along the channel dimension. While retaining adaptive layer normalization, this approach uses only the timestep embedding to predict the scale and shift within blocks. However, we see that in-context conditioning is less effective for depth image conditioning, as it requires expanding the lower-dimensional SMPL parameters to match the higher dimensionality of the depth images before concatenation, which reduces the efficiency of feature integration.

\noindent \textbf{Design Choice 3} - With Attention Block vs. Without Attention Block — We assess the impact of incorporating attention blocks in the final layers of the diffusion model. The results indicate that adding attention blocks noticeably improves in-bed pose prediction performance.

\section{Conclusion}
\label{sec:conclusion}

In this work, we present a diffusion-based framework for in-bed human mesh recovery, designed to enhance generalization and accuracy in healthcare environments with limited real-world data. By leveraging synthetic data and a Sim-to-Real Transfer Framework, our approach effectively addresses challenges posed by privacy concerns, occlusions, and data scarcity. Extensive experiments demonstrated the robustness of our method across varying covering conditions and high real-to-simulation ratios, consistently outperforming existing methods in MPJPE and PVE metrics. Additionally, our model's adaptability across different environments and reduced reliance on real-world data offer an efficient and scalable solution for clinical deployment.

\clearpage
{
    \small
    \bibliographystyle{ieeenat_fullname}
    \bibliography{main}

\begin{thebibliography}{49}
\providecommand{\natexlab}[1]{#1}
\providecommand{\url}[1]{\texttt{#1}}
\expandafter\ifx\csname urlstyle\endcsname\relax
  \providecommand{\doi}[1]{doi: #1}\else
  \providecommand{\doi}{doi: \begingroup \urlstyle{rm}\Url}\fi

\bibitem[Black et~al.(2023)Black, Patel, Tesch, and Yang]{Black_CVPR_2023}
Michael~J. Black, Priyanka Patel, Joachim Tesch, and Jinlong Yang.
\newblock {BEDLAM}: A synthetic dataset of bodies exhibiting detailed lifelike animated motion.
\newblock In \emph{Proceedings IEEE/CVF Conf.~on Computer Vision and Pattern Recognition (CVPR)}, pages 8726--8737, 2023.

\bibitem[Cai et~al.(2024)Cai, Hu, Hou, Yao, and Huang]{cai2024disentangled}
Qingyuan Cai, Xuecai Hu, Saihui Hou, Li Yao, and Yongzhen Huang.
\newblock Disentangled diffusion-based 3d human pose estimation with hierarchical spatial and temporal denoiser.
\newblock In \emph{Proceedings of the AAAI Conference on Artificial Intelligence}, pages 882--890, 2024.

\bibitem[Choi et~al.(2021)Choi, Kim, Jeong, Gwon, and Yoon]{choi2021ilvr}
Jooyoung Choi, Sungwon Kim, Yonghyun Jeong, Youngjune Gwon, and Sungroh Yoon.
\newblock Ilvr: Conditioning method for denoising diffusion probabilistic models.
\newblock \emph{arXiv preprint arXiv:2108.02938}, 2021.

\bibitem[Clever et~al.(2020)Clever, Erickson, Kapusta, Turk, Liu, and Kemp]{clever2020bodies}
Henry~M Clever, Zackory Erickson, Ariel Kapusta, Greg Turk, Karen Liu, and Charles~C Kemp.
\newblock Bodies at rest: 3d human pose and shape estimation from a pressure image using synthetic data.
\newblock In \emph{Proceedings of the IEEE/CVF conference on computer vision and pattern recognition}, pages 6215--6224, 2020.

\bibitem[Clever et~al.(2022)Clever, Grady, Turk, and Kemp]{clever2022bodypressure}
Henry~M Clever, Patrick~L Grady, Greg Turk, and Charles~C Kemp.
\newblock Bodypressure-inferring body pose and contact pressure from a depth image.
\newblock \emph{IEEE Transactions on Pattern Analysis and Machine Intelligence}, 45\penalty0 (1):\penalty0 137--153, 2022.

\bibitem[Dwivedi et~al.(2024)Dwivedi, Sun, Patel, Feng, and Black]{dwivedi2024tokenhmr}
Sai~Kumar Dwivedi, Yu Sun, Priyanka Patel, Yao Feng, and Michael~J Black.
\newblock Tokenhmr: Advancing human mesh recovery with a tokenized pose representation.
\newblock In \emph{Proceedings of the IEEE/CVF Conference on Computer Vision and Pattern Recognition}, pages 1323--1333, 2024.

\bibitem[Foo et~al.(2023)Foo, Gong, Rahmani, and Liu]{foo2023distribution}
Lin~Geng Foo, Jia Gong, Hossein Rahmani, and Jun Liu.
\newblock Distribution-aligned diffusion for human mesh recovery.
\newblock In \emph{Proceedings of the IEEE/CVF International Conference on Computer Vision}, pages 9221--9232, 2023.

\bibitem[Goel et~al.(2023)Goel, Pavlakos, Rajasegaran, Kanazawa*, and Malik*]{goel2023humans}
Shubham Goel, Georgios Pavlakos, Jathushan Rajasegaran, Angjoo Kanazawa*, and Jitendra Malik*.
\newblock Humans in 4{D}: Reconstructing and tracking humans with transformers.
\newblock In \emph{International Conference on Computer Vision (ICCV)}, 2023.

\bibitem[Gong et~al.(2023)Gong, Foo, Fan, Ke, Rahmani, and Liu]{gong2023diffpose}
Jia Gong, Lin~Geng Foo, Zhipeng Fan, Qiuhong Ke, Hossein Rahmani, and Jun Liu.
\newblock Diffpose: Toward more reliable 3d pose estimation.
\newblock In \emph{Proceedings of the IEEE/CVF Conference on Computer Vision and Pattern Recognition}, pages 13041--13051, 2023.

\bibitem[He et~al.(2016)He, Zhang, Ren, and Sun]{he2016deep}
Kaiming He, Xiangyu Zhang, Shaoqing Ren, and Jian Sun.
\newblock Deep residual learning for image recognition.
\newblock In \emph{Proceedings of the IEEE conference on computer vision and pattern recognition}, pages 770--778, 2016.

\bibitem[Ho et~al.(2020)Ho, Jain, and Abbeel]{ho2020denoising}
Jonathan Ho, Ajay Jain, and Pieter Abbeel.
\newblock Denoising diffusion probabilistic models.
\newblock \emph{Advances in Neural Information Processing Systems}, 33:\penalty0 6840--6851, 2020.

\bibitem[Kanazawa et~al.(2018)Kanazawa, Black, Jacobs, and Malik]{kanazawaHMR18}
Angjoo Kanazawa, Michael~J. Black, David~W. Jacobs, and Jitendra Malik.
\newblock End-to-end recovery of human shape and pose.
\newblock In \emph{CVPR}, 2018.

\bibitem[Kolotouros et~al.(2019)Kolotouros, Pavlakos, Black, and Daniilidis]{Kolotouros2019SPIN}
Nikos Kolotouros, Georgios Pavlakos, Michael~J Black, and Kostas Daniilidis.
\newblock Learning to reconstruct 3d human pose and shape via model-fitting in the loop.
\newblock In \emph{Proceedings of the IEEE/CVF International Conference on Computer Vision}, pages 2252--2261, 2019.

\bibitem[Li et~al.(2021)Li, Xu, Chen, Bian, Yang, and Lu]{li2021hybrik}
Jiefeng Li, Chao Xu, Zhicun Chen, Siyuan Bian, Lixin Yang, and Cewu Lu.
\newblock Hybrik: A hybrid analytical-neural inverse kinematics solution for 3d human pose and shape estimation.
\newblock In \emph{Proceedings of the IEEE/CVF Conference on Computer Vision and Pattern Recognition}, pages 3383--3393, 2021.

\bibitem[Li et~al.(2023{\natexlab{a}})Li, Bian, Liu, Tang, Wang, and Lu]{li2023niki}
Jiefeng Li, Siyuan Bian, Qi Liu, Jiasheng Tang, Fan Wang, and Cewu Lu.
\newblock Niki: Neural inverse kinematics with invertible neural networks for 3d human pose and shape estimation.
\newblock In \emph{Proceedings of the IEEE/CVF Conference on Computer Vision and Pattern Recognition}, pages 12933--12942, 2023{\natexlab{a}}.

\bibitem[Li et~al.(2023{\natexlab{b}})Li, Bian, Xu, Chen, Yang, and Lu]{li2023hybrik}
Jiefeng Li, Siyuan Bian, Chao Xu, Zhicun Chen, Lixin Yang, and Cewu Lu.
\newblock Hybrik-x: Hybrid analytical-neural inverse kinematics for whole-body mesh recovery.
\newblock \emph{arXiv preprint arXiv:2304.05690}, 2023{\natexlab{b}}.

\bibitem[Li et~al.(2023{\natexlab{c}})Li, Liu, and Wu]{li2023ego}
Jiaman Li, Karen Liu, and Jiajun Wu.
\newblock Ego-body pose estimation via ego-head pose estimation.
\newblock In \emph{Proceedings of the IEEE/CVF Conference on Computer Vision and Pattern Recognition}, pages 17142--17151, 2023{\natexlab{c}}.

\bibitem[Li et~al.(2025)Li, Yang, Kuang, Wu, Wang, Xiao, and Chen]{li2025controlnetpp}
Ming Li, Taojiannan Yang, Huafeng Kuang, Jie Wu, Zhaoning Wang, Xuefeng Xiao, and Chen Chen.
\newblock Controlnet++: Improving conditional controls with efficient consistency feedback.
\newblock In \emph{European Conference on Computer Vision}, pages 129--147. Springer, 2025.

\bibitem[Li et~al.(2022)Li, Liu, Zhang, Xu, and Yan]{li2022cliff}
Zhihao Li, Jianzhuang Liu, Zhensong Zhang, Songcen Xu, and Youliang Yan.
\newblock Cliff: Carrying location information in full frames into human pose and shape estimation.
\newblock In \emph{European Conference on Computer Vision}, pages 590--606. Springer, 2022.

\bibitem[Lin et~al.(2023)Lin, Zeng, Wang, Zhang, and Li]{lin2023one}
Jing Lin, Ailing Zeng, Haoqian Wang, Lei Zhang, and Yu Li.
\newblock One-stage 3d whole-body mesh recovery with component aware transformer.
\newblock In \emph{Proceedings of the IEEE/CVF Conference on Computer Vision and Pattern Recognition}, pages 21159--21168, 2023.

\bibitem[Lin et~al.(2021)Lin, Wang, and Liu]{lin2021metro}
Kevin Lin, Lijuan Wang, and Zicheng Liu.
\newblock End-to-end human pose and mesh reconstruction with transformers.
\newblock In \emph{Proceedings of the IEEE/CVF Conference on Computer Vision and Pattern Recognition}, pages 1954--1963, 2021.

\bibitem[Liu et~al.(2023)Liu, Huang, Fu, Li, Su, and Ostadabbas]{slpdata}
Shuangjun Liu, Xiaofei Huang, Nihang Fu, Cheng Li, Zhongnan Su, and Sarah Ostadabbas.
\newblock Simultaneously-collected multimodal lying pose dataset: Enabling in-bed human pose monitoring.
\newblock \emph{IEEE Transactions on Pattern Analysis and Machine Intelligence}, 45\penalty0 (1):\penalty0 1106--1118, 2023.

\bibitem[Loper et~al.(2015)Loper, Mahmood, Romero, Pons-Moll, and Black]{SMPL:2015}
Matthew Loper, Naureen Mahmood, Javier Romero, Gerard Pons-Moll, and Michael~J. Black.
\newblock {SMPL}: A skinned multi-person linear model.
\newblock \emph{ACM TOG}, 2015.

\bibitem[Loshchilov(2017)]{loshchilov2017decoupled}
I Loshchilov.
\newblock Decoupled weight decay regularization.
\newblock \emph{arXiv preprint arXiv:1711.05101}, 2017.

\bibitem[Lu et~al.(2023)Lu, Lin, Dou, Zhang, Deng, and Wang]{lu2023dposer}
Junzhe Lu, Jing Lin, Hongkun Dou, Yulun Zhang, Yue Deng, and Haoqian Wang.
\newblock Dposer: Diffusion model as robust 3d human pose prior.
\newblock \emph{arXiv preprint arXiv:2312.05541}, 2023.

\bibitem[Mehta et~al.(2017)Mehta, Sridhar, Sotnychenko, Rhodin, Shafiei, Seidel, Xu, Casas, and Theobalt]{mehta2017vnect}
Dushyant Mehta, Srinath Sridhar, Oleksandr Sotnychenko, Helge Rhodin, Mohammad Shafiei, Hans-Peter Seidel, Weipeng Xu, Dan Casas, and Christian Theobalt.
\newblock Vnect: Real-time 3d human pose estimation with a single rgb camera.
\newblock \emph{Acm transactions on graphics (tog)}, 36\penalty0 (4):\penalty0 1--14, 2017.

\bibitem[Nichol and Dhariwal(2021)]{nichol2021improved}
Alexander~Quinn Nichol and Prafulla Dhariwal.
\newblock Improved denoising diffusion probabilistic models.
\newblock In \emph{International conference on machine learning}, pages 8162--8171. PMLR, 2021.

\bibitem[Patel et~al.(2021)Patel, Huang, Tesch, Hoffmann, Tripathi, and Black]{Patel:CVPR:2021}
Priyanka Patel, Chun-Hao~P. Huang, Joachim Tesch, David~T. Hoffmann, Shashank Tripathi, and Michael~J. Black.
\newblock {AGORA}: Avatars in geography optimized for regression analysis.
\newblock In \emph{Proceedings IEEE/CVF Conf.~on Computer Vision and Pattern Recognition ({CVPR})}, 2021.

\bibitem[Pavllo et~al.(2019)Pavllo, Feichtenhofer, Grangier, and Auli]{pavllo20193d}
Dario Pavllo, Christoph Feichtenhofer, David Grangier, and Michael Auli.
\newblock 3d human pose estimation in video with temporal convolutions and semi-supervised training.
\newblock In \emph{Proceedings of the IEEE/CVF conference on computer vision and pattern recognition}, pages 7753--7762, 2019.

\bibitem[Peebles and Xie(2023)]{peebles2023scalable}
William Peebles and Saining Xie.
\newblock Scalable diffusion models with transformers.
\newblock In \emph{Proceedings of the IEEE/CVF International Conference on Computer Vision}, pages 4195--4205, 2023.

\bibitem[Rombach et~al.(2022)Rombach, Blattmann, Lorenz, Esser, and Ommer]{rombach2022high}
Robin Rombach, Andreas Blattmann, Dominik Lorenz, Patrick Esser, and Bj{\"o}rn Ommer.
\newblock High-resolution image synthesis with latent diffusion models.
\newblock In \emph{Proceedings of the IEEE/CVF conference on computer vision and pattern recognition}, pages 10684--10695, 2022.

\bibitem[Saharia et~al.(2022)Saharia, Chan, Saxena, Li, Whang, Denton, Ghasemipour, Gontijo-Lopes, Ayan, Salimans, Ho, Fleet, and Norouzi]{photorealistic}
Chitwan Saharia, William Chan, Saurabh Saxena, Lala Li, Jay Whang, Emily Denton, Seyed Kamyar~Seyed Ghasemipour, Raphael Gontijo-Lopes, Burcu~Karagol Ayan, Tim Salimans, Jonathan Ho, David~J. Fleet, and Mohammad Norouzi.
\newblock Photorealistic text-to-image diffusion models with deep language understanding.
\newblock In \emph{Advances in Neural Information Processing Systems}, 2022.

\bibitem[Shan et~al.(2023)Shan, Liu, Zhang, Wang, Han, Wang, Ma, and Gao]{shan2023diffusion}
Wenkang Shan, Zhenhua Liu, Xinfeng Zhang, Zhao Wang, Kai Han, Shanshe Wang, Siwei Ma, and Wen Gao.
\newblock Diffusion-based 3d human pose estimation with multi-hypothesis aggregation.
\newblock \emph{arXiv preprint arXiv:2303.11579}, 2023.

\bibitem[Song et~al.(2020)Song, Meng, and Ermon]{song2020denoising}
Jiaming Song, Chenlin Meng, and Stefano Ermon.
\newblock Denoising diffusion implicit models.
\newblock \emph{arXiv preprint arXiv:2010.02502}, 2020.

\bibitem[Song et~al.(2021)Song, Sohl-Dickstein, Kingma, Kumar, Ermon, and Poole]{song2021scorebased}
Yang Song, Jascha Sohl-Dickstein, Diederik~P Kingma, Abhishek Kumar, Stefano Ermon, and Ben Poole.
\newblock Score-based generative modeling through stochastic differential equations.
\newblock In \emph{International Conference on Learning Representations}, 2021.

\bibitem[Stathopoulos et~al.(2024)Stathopoulos, Han, and Metaxas]{stathopoulos2024scorehmr}
Anastasis Stathopoulos, Ligong Han, and Dimitris Metaxas.
\newblock Score-guided diffusion for 3d human recovery.
\newblock In \emph{Proceedings of the IEEE/CVF Conference on Computer Vision and Pattern Recognition}, pages 906--915, 2024.

\bibitem[Ta et~al.(2024)Ta, Dutta, Kundu, Lal, Cruz, Raychaudhuri, and Roy-Chowdhury]{ta2024MOPED}
Calvin-Khang Ta, Arindam Dutta, Rohit Kundu, Rohit Lal, Hannah~Dela Cruz, Dripta~S Raychaudhuri, and Amit Roy-Chowdhury.
\newblock Multi-modal pose diffuser: A multimodal generative conditional pose prior.
\newblock \emph{arXiv preprint arXiv:2410.14540}, 2024.

\bibitem[Tandon et~al.(2024)Tandon, Goyal, Clever, and Erickson]{tandon2024bodymap}
Abhishek Tandon, Anujraaj Goyal, Henry~M Clever, and Zackory Erickson.
\newblock Bodymap-jointly predicting body mesh and 3d applied pressure map for people in bed.
\newblock In \emph{Proceedings of the IEEE/CVF Conference on Computer Vision and Pattern Recognition}, pages 2480--2489, 2024.

\bibitem[Tian et~al.(2023)Tian, Zhang, Liu, and Wang]{tian2023recovering}
Yating Tian, Hongwen Zhang, Yebin Liu, and Limin Wang.
\newblock Recovering 3d human mesh from monocular images: A survey.
\newblock \emph{IEEE transactions on pattern analysis and machine intelligence}, 2023.

\bibitem[Wang and Daniilidis(2023)]{wang2023refit}
Yufu Wang and Kostas Daniilidis.
\newblock Refit: Recurrent fitting network for 3d human recovery.
\newblock In \emph{Proceedings of the IEEE/CVF International Conference on Computer Vision}, pages 14644--14654, 2023.

\bibitem[Yang et~al.(2023)Yang, Cai, Mei, Liu, Chen, Xiao, Wei, Qing, Wei, Dai, Wu, Qian, Lin, Liu, and Yang]{yang2023synbody}
Zhitao Yang, Zhongang Cai, Haiyi Mei, Shuai Liu, Zhaoxi Chen, Weiye Xiao, Yukun Wei, Zhongfei Qing, Chen Wei, Bo Dai, Wayne Wu, Chen Qian, Dahua Lin, Ziwei Liu, and Lei Yang.
\newblock Synbody: Synthetic dataset with layered human models for 3d human perception and modeling.
\newblock In \emph{Proceedings of the IEEE/CVF International Conference on Computer Vision (ICCV)}, pages 20282--20292, 2023.

\bibitem[Yin et~al.(2022)Yin, Robinson, and Fu]{yin2022multimodal}
Yu Yin, Joseph~P Robinson, and Yun Fu.
\newblock Multimodal in-bed pose and shape estimation under the blankets.
\newblock In \emph{Proceedings of the 30th ACM International Conference on Multimedia}, pages 2411--2419, 2022.

\bibitem[Zhang et~al.(2021)Zhang, Tian, Zhou, Ouyang, Liu, Wang, and Sun]{pymaf2021}
Hongwen Zhang, Yating Tian, Xinchi Zhou, Wanli Ouyang, Yebin Liu, Limin Wang, and Zhenan Sun.
\newblock Pymaf: 3d human pose and shape regression with pyramidal mesh alignment feedback loop.
\newblock In \emph{Proceedings of the IEEE International Conference on Computer Vision}, 2021.

\bibitem[Zhang et~al.(2023)Zhang, Rao, and Agrawala]{zhang2023controlnet}
Lvmin Zhang, Anyi Rao, and Maneesh Agrawala.
\newblock Adding conditional control to text-to-image diffusion models.
\newblock In \emph{Proceedings of the IEEE/CVF International Conference on Computer Vision}, pages 3836--3847, 2023.

\bibitem[Zhang et~al.(2024)Zhang, Wu, Wang, Luo, Zhang, Zhao, Vajda, Metaxas, and Yu]{zhang2024avid}
Zhixing Zhang, Bichen Wu, Xiaoyan Wang, Yaqiao Luo, Luxin Zhang, Yinan Zhao, Peter Vajda, Dimitris Metaxas, and Licheng Yu.
\newblock Avid: Any-length video inpainting with diffusion model.
\newblock In \emph{Proceedings of the IEEE/CVF Conference on Computer Vision and Pattern Recognition}, pages 7162--7172, 2024.

\bibitem[Zhao et~al.(2024)Zhao, Zheng, Liu, and Chen]{zhao2024single}
Qitao Zhao, Ce Zheng, Mengyuan Liu, and Chen Chen.
\newblock A single 2d pose with context is worth hundreds for 3d human pose estimation.
\newblock \emph{Advances in Neural Information Processing Systems}, 36, 2024.

\bibitem[Zheng et~al.(2023)Zheng, Mendieta, Yang, Qi, and Chen]{zheng2023feater}
Ce Zheng, Matias Mendieta, Taojiannan Yang, Guo-Jun Qi, and Chen Chen.
\newblock Feater: An efficient network for human reconstruction via feature map-based transformer.
\newblock In \emph{IEEE/CVF Conference on Computer Vision and Pattern Recognition (CVPR)}, 2023.

\bibitem[Zheng et~al.(2025)Zheng, Liu, Peng, Wu, Wang, and Chen]{zheng2023diffmesh}
Ce Zheng, Xianpeng Liu, Qucheng Peng, Tianfu Wu, Pu Wang, and Chen Chen.
\newblock Diffmesh: A motion-aware diffusion framework for human mesh recovery from videos.
\newblock In \emph{Proceedings of the Winter Conference on Applications of Computer Vision (WACV)}, 2025.

\bibitem[Zhu et~al.(2023)Zhu, Ma, Liu, Liu, Wu, and Wang]{zhu2023motionbert}
Wentao Zhu, Xiaoxuan Ma, Zhaoyang Liu, Libin Liu, Wayne Wu, and Yizhou Wang.
\newblock Motionbert: A unified perspective on learning human motion representations.
\newblock In \emph{Proceedings of the IEEE/CVF International Conference on Computer Vision}, pages 15085--15099, 2023.

\end{thebibliography}
}
\clearpage
\setcounter{page}{1}
\maketitlesupplementary

In the supplementary material, we provide additional discussions on synthetic datasets for human mesh recovery(Sec.~\ref{sec:discussion_syn}), as well as additional details on data augmentation (Sec.~\ref{sec:data_aug}), loss functions (Sec.~\ref{sec:sup_imp}), ablation studies (Sec.~\ref{sec:sup_ablation}), and visualization examples (Sec.~\ref{sec:sup_viz}) to further demonstrate the effectiveness of the DiSRT-In-Bed framework.

\section{Synthetic Datasets for Human Mesh Recovery}
\label{sec:discussion_syn}
Synthetic datasets are widely used in advancing 3D human mesh recovery by providing large-scale, diverse, and accurately labeled data that would be difficult and expensive to obtain through real-world capture. Prior works such as AGORA~\cite{Patel:CVPR:2021}, BEDLAM~\cite{Black_CVPR_2023}, and SynBody~\cite{yang2023synbody} demonstrate that incorporating synthetic data in training enhances human mesh recovery performance. However, general synthetic datasets are not directly applicable to in-bed scenarios, as lying poses are underrepresented. For in-bed human mesh recovery, BodyPressure~\cite{clever2022bodypressure} builds upon Bodies at Rest~\cite{clever2020bodies} to enhance synthetic dataset generation. It leverages physics-based simulation to produce realistic depth and pressure images, better capturing human-bed interactions and occlusions. Additionally, BodyPressure and BodyMAP further demonstrate that scenario-specific synthetic datasets can improve in-bed human mesh estimation.

\section{Additional Details about Training Strategy}

\subsection{Data Augmentation}
\label{sec:data_aug}
\begin{figure}[ht]
  \centering
   \includegraphics[width=\linewidth]{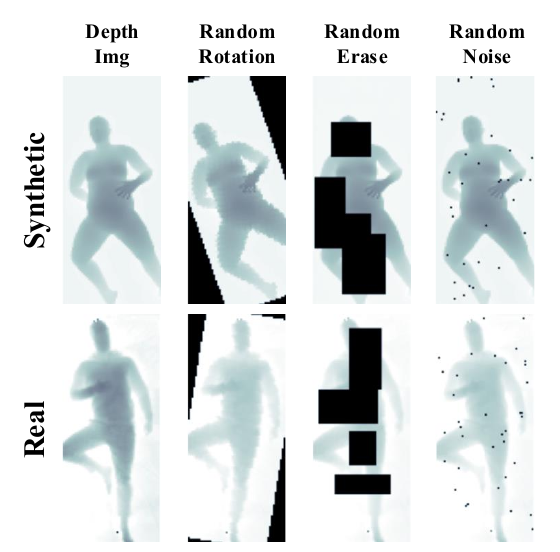}
   \caption{\textbf{Illustration of Data Augmentation Operations.}}
   \label{fig:data_aug}
\end{figure}

To enhance the robustness of the diffusion model during training and fine-tuning, we apply various data augmentation techniques to the input depth images for both synthetic and real datasets, simulating complex real-world scenarios. As shown in Fig.~\ref{fig:data_aug}, the following augmentations are applied:
\begin{itemize}
    \item \textbf{Random Rotation:} Depth images are randomly rotated to introduce variability in human in-bed poses.
    \item \textbf{Random Erase:} Portions of the depth image are randomly masked, simulating occlusions caused by objects such as tables or blankets covering parts of the human body.
    \item \textbf{Random Noise:} Gaussian noise is added to mimic the noise introduced by depth sensors and environmental factors.
\end{itemize}

\noindent These augmentations aim to improve the model's ability to generalize to diverse and challenging real-world conditions.

\subsection{Loss Functions}
\label{sec:sup_imp}
The total diffusion loss used to train and fine-tune the diffusion model contains two components: SMPL parameter loss and vertex position loss. For SMPL parameter loss, we employ standard human pose and shape regularization loss utilized in BodyMAP~\cite{tandon2024bodymap} as follows:
\begin{equation}
\small
\begin{aligned}
\mathcal{L}_{\text{SMPL}} &= \lambda_{\boldsymbol{\beta}} \| \boldsymbol{\beta} - \hat{\boldsymbol{\beta}} \|_1 
+ \lambda_{\boldsymbol{\theta}} \| \boldsymbol{\theta} - \hat{\boldsymbol{\theta}} \|_1 \\
&\quad + \lambda_{\boldsymbol{\psi_x}} \left( \| \mathbf{u} - \hat{\mathbf{u}} \|_1 + \| \mathbf{v} - \hat{\mathbf{v}} \|_1 \right) 
+ \lambda_{\boldsymbol{J}}  \sum_{i=1}^{24}\| \mathbf{j}_i - \hat{\mathbf{j}}_i \|_2, \\
\lambda_{\boldsymbol{\beta}} &= \frac{1}{10 \sigma_{\boldsymbol{\beta}}}, \quad
\lambda_{\boldsymbol{\theta}} = \frac{1}{69 \sigma_{\boldsymbol{\theta}}}, \quad
\lambda_{\boldsymbol{\psi_x}} = \frac{1}{6 \sigma_{\psi_x}}, \quad
\lambda_{\boldsymbol{J}} = \frac{1}{24 \sigma_J},
\end{aligned}
\end{equation}

where each hyper-parameter term is normalized by standard deviations of body parameters $ \sigma_{\boldsymbol{\beta}}$, joint angles $\sigma_{\boldsymbol{\theta}}$, continuous global rotation $\sigma_{\boldsymbol{\psi_x}}$ and Cartesian joint positions, computed from the entire synthetic training set. $ \mathbf{j}_i \subset \mathbf{J} $ represents the Cartesian position of a single joint.
Additionally, we use vertex loss to further enhance diffusion stability and performance:
\begin{equation}
\mathcal{L}_{\text{v2v}} = \frac{1}{N_{\mathbf{V}} \sigma_{\mathbf{V}}} \sum_{i=1}^{N_{\mathbf{V}}} \| \mathbf{v}_i - \hat{\mathbf{v}}_i \|_2
\end{equation}
where $\mathbf{v}_i \subset \mathbf{V} $ represents the Cartesian position of a single human mesh vertex, $ N_{\mathbf{V}} = 6890 $ vertices, and the loss term is normalized by $ \sigma_{\mathbf{V}} $.

Thus, the total loss for the diffusion reverse process is:
\begin{equation}
    \mathcal{L_{\text{total}}} = \mathcal{L}_{\text{SMPL}} + \lambda_{\text{v2v}} \mathcal{L}_{\text{v2v}},
\end{equation}
where $\lambda_{\text{v2v}}$ is a tunable hyper-parameters. We set $\lambda_\text{v2v} = 1.0$ for all the experiments.

\subsection{Learning Rate Scheduler}
As mentioned in Sec.~4.3, we adopt a linearly adjusted learning rate scheduler to adapt to varying amount of real-world data during the fine-tuning stage. Specifically, given the initial learning rate \(\text{lr\_init}\), the current step index \(\text{step\_cur}\), and the total number of fine-tuning steps \(\text{steps\_total}\), the current learning rate is computed as:  
\begin{equation}
    \text{lr\_cur} = \left(1 - \frac{\text{step\_cur}}{\text{steps\_total} + 1} \right) \text{lr\_init}.
\end{equation}

\section{Additional Ablation Study}
\label{sec:sup_ablation}

\begin{figure*}[ht]
  \centering
  \subcaptionbox{Ablation on Synthetic Data\label{fig:ablation_pve_syn}}{%
    \includegraphics[width=0.32\linewidth]{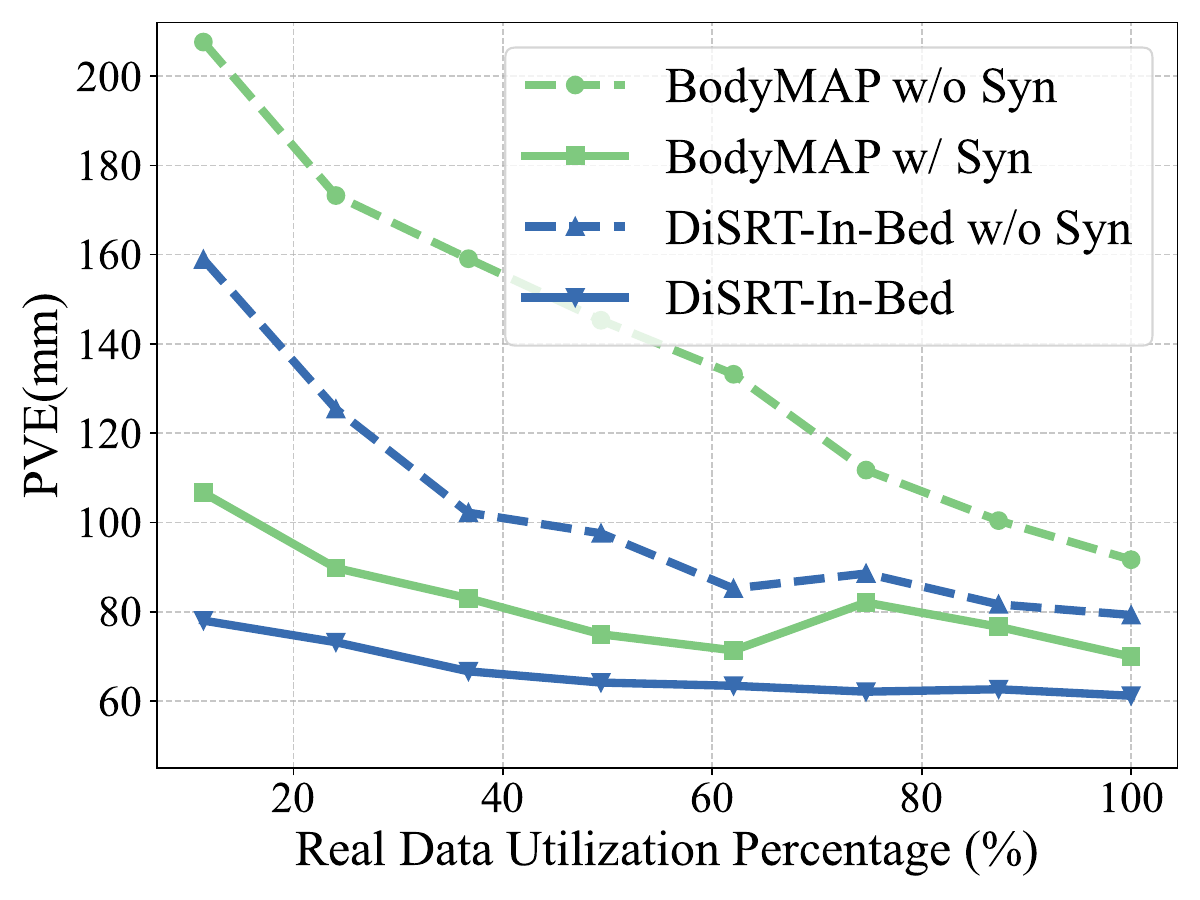}
  }
  \subcaptionbox{Ablation on Training Strategies\label{fig:ablation_pve_s2r}}{%
    \includegraphics[width=0.32\linewidth]{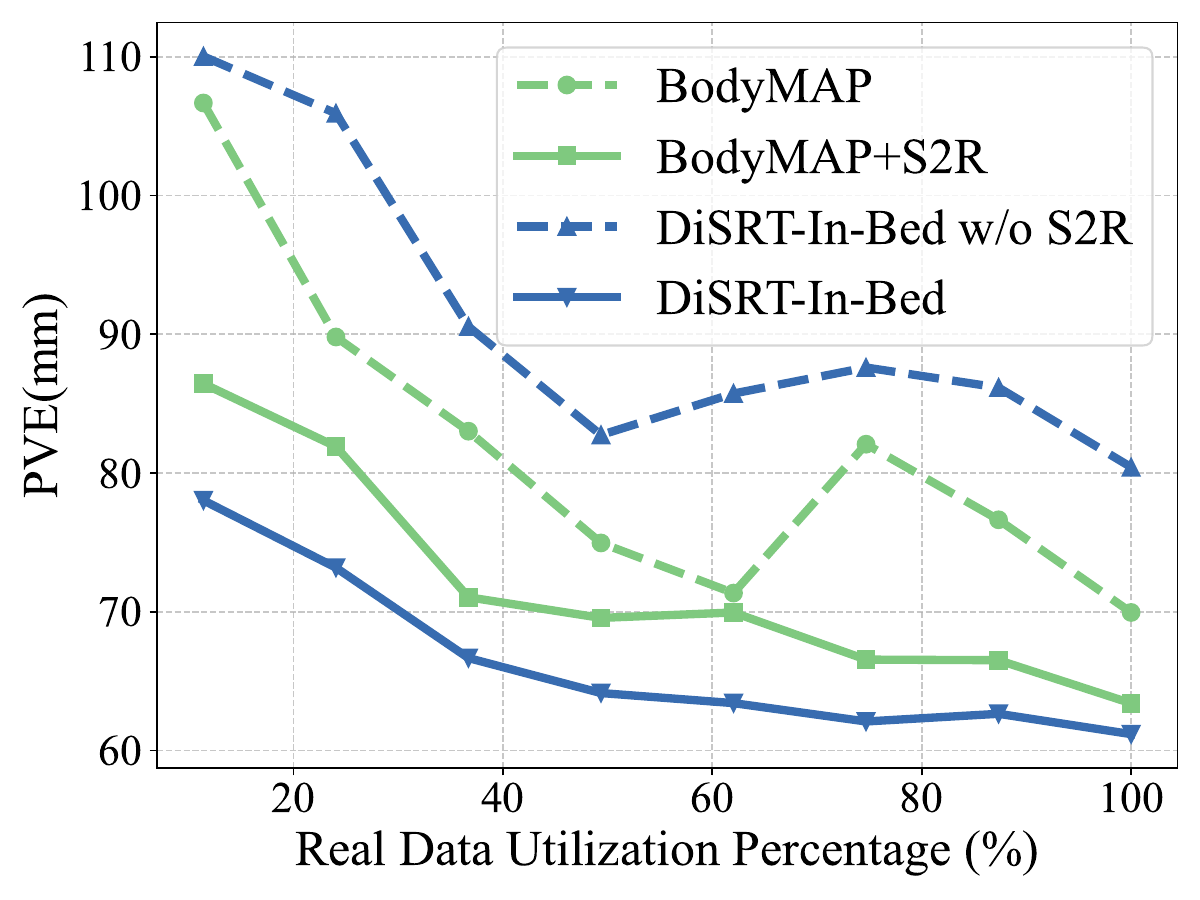}
  }
  \subcaptionbox{Ablation on Different Architectures\label{fig:ablation_pve_arch}}{%
    \includegraphics[width=0.32\linewidth]{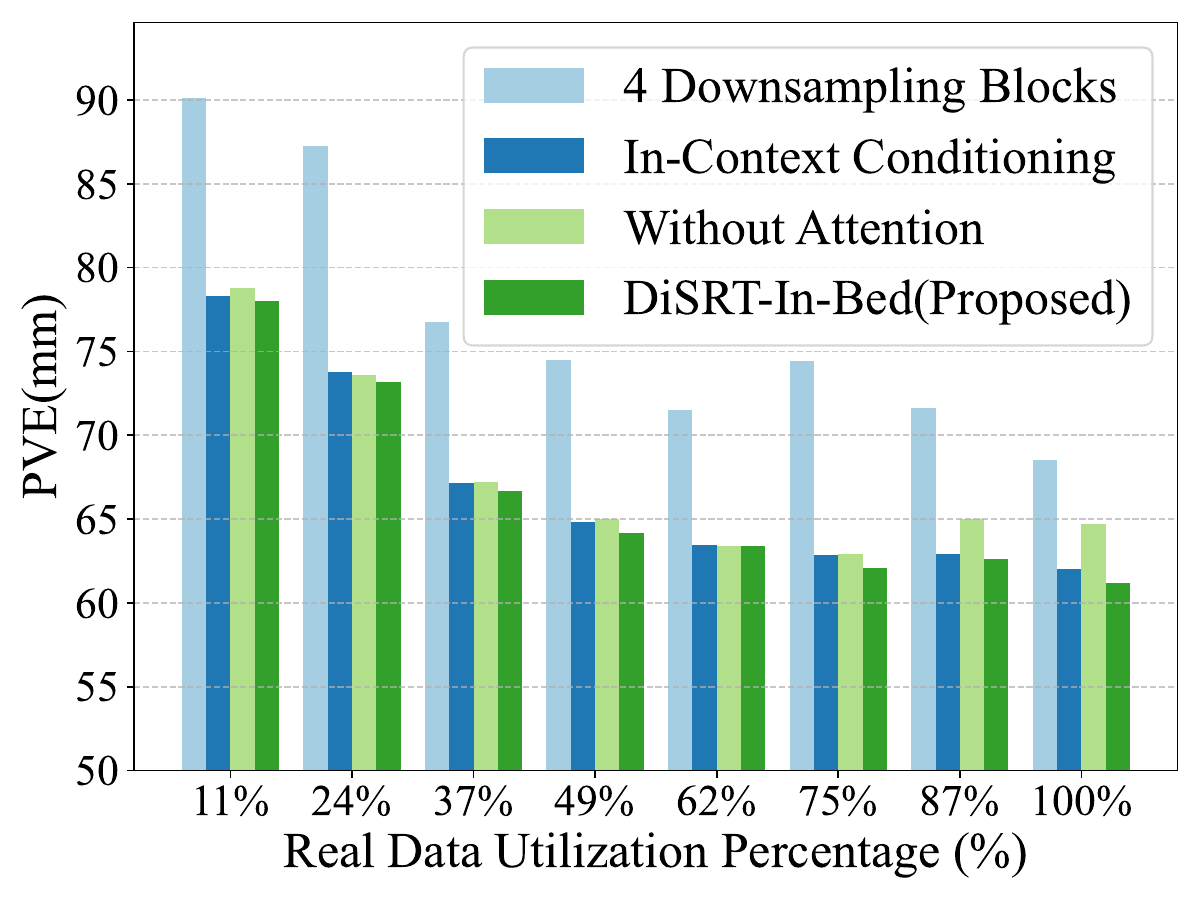}
  }
  \caption{\textbf{Additional Ablation Study on Diffusion-Based Sim-to-Real Transfer Framework.}}
  \label{fig:ablation}
\end{figure*}
\subsection{Effectiveness of Loss function}

\begin{table}[h]
  \centering
  \begin{tabular}{lcc}
    \toprule
    Loss & MJPJE & PVE \\
    \midrule
    SMPL Loss & 53.48         & 66.86\\
    SMPL Loss + v2v Loss & 50.81         & 61.18\\
    \bottomrule
  \end{tabular}
  \caption{\textbf{Ablation on Loss Function.}}
  \label{tab:loss_ablation}
  \vspace{-5pt}
\end{table}

We conduct an ablation study by comparing models trained with different loss functions using the complete synthetic and real training datasets. Tab.~\ref{tab:loss_ablation} shows that adding the v2v loss term to the total loss function enhances the model's performance in mesh estimation in terms of both MPJPE and PVE metrics.

\subsection{Additional Comparisons of PVE Results}
In addition to the results presented in Sec.5.5 of the main paper, we provide charts for the PVE metric to further demonstrate the effectiveness of our Sim-to-Real Transfer Framework and the proposed diffusion model architecture. The trends observed in PVE results across varying real data utilization percentages align closely with those of the MPJPE results.

Fig.~\ref{fig:ablation_pve_syn} shows that leveraging synthetic data substantially enhances model performance in the PVE metric. Fig.~\ref{fig:ablation_pve_s2r} demonstrates that integrating our Sim-to-Real Transfer Framework into the BodyMAP model results in significant improvements, particularly under scenarios with limited access to real-world data. Additionally, Fig.~\ref{fig:ablation_pve_arch} compares four diffusion model designs on the PVE metric. Although the differences in PVE are less pronounced compared to the MPJPE results shown in Fig.5c of the main paper, our proposed architecture consistently outperforms other design choices.

\subsection{Effectiveness of Fine-tuning Strategies}
\begin{figure}[ht]
  \centering
    \includegraphics[width=0.8\linewidth]{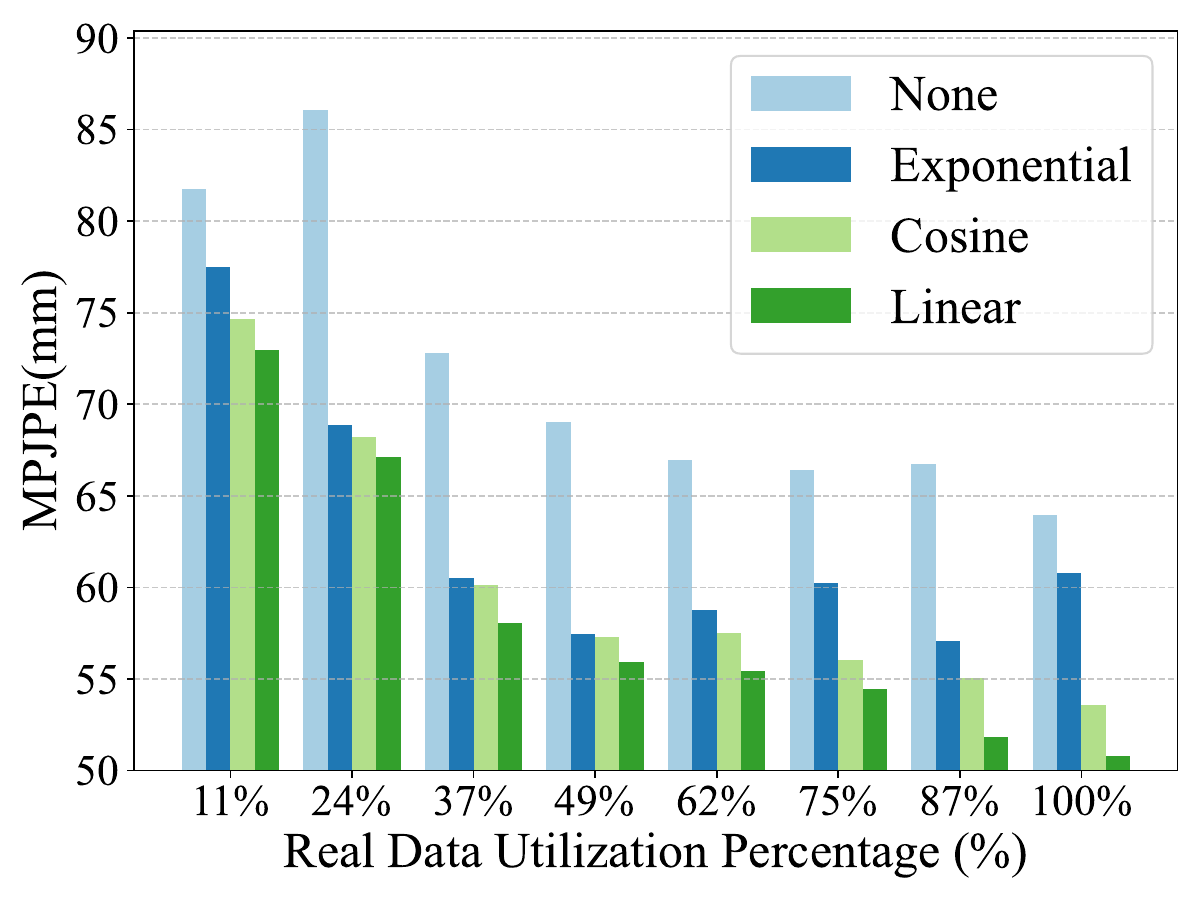}
    \includegraphics[width=0.8\linewidth]{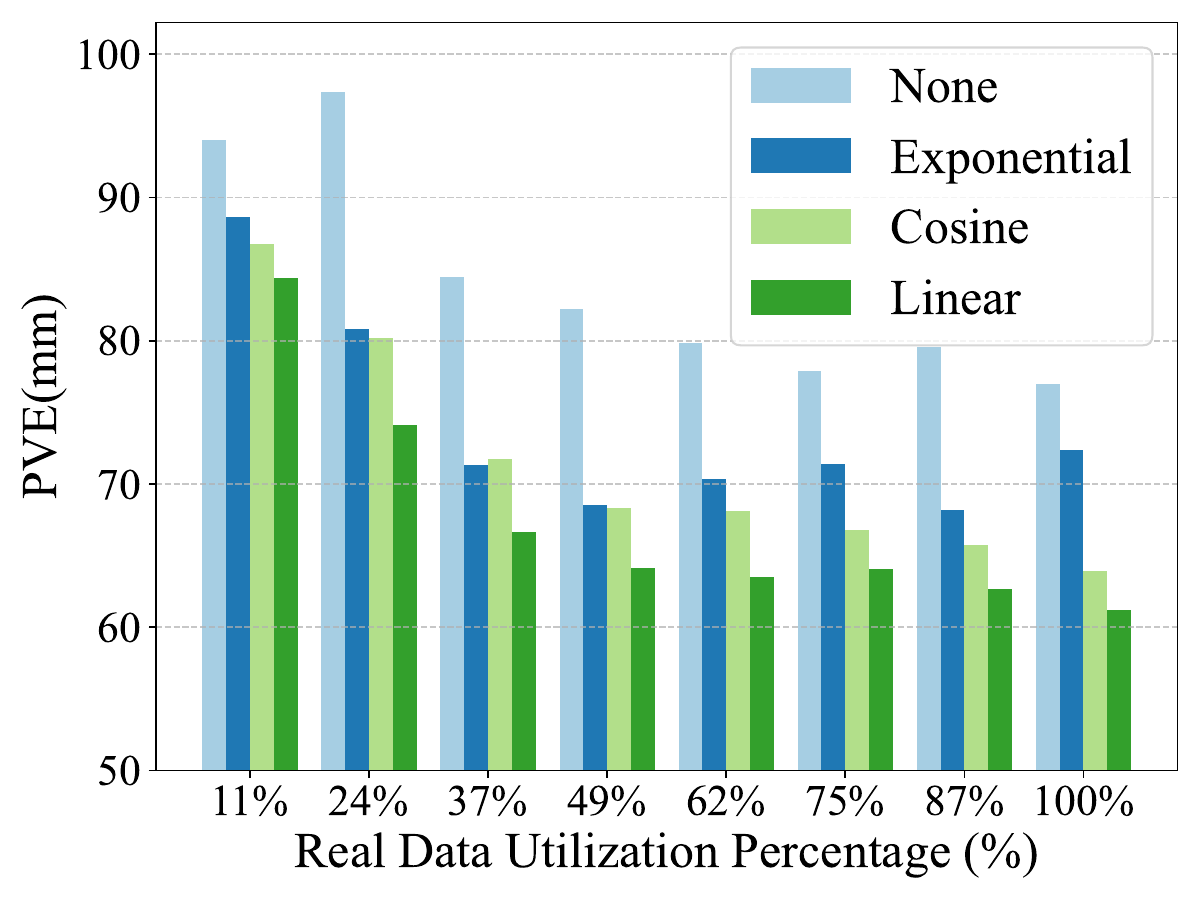}
  \caption{\textbf{Ablation Study on Fine-tuning Schedulers.}}
  \label{fig:ablation_scheduler}
\end{figure}

In the fine-tuning stage, we introduce a linearly and automatically adjusted scheduler as described in Sec.4.3 of the main paper. 
The initial learning rate is set to match that of the training stage, i.e., $lr = 1 \times 10^{-4}$. During fine-tuning, the learning rate and weight decay are updated at each diffusion step using the AdamW optimizer. 
Specifically, for each step, we input a batch of depth images paired with randomly generated timesteps and generate noisy SMPL parameters by iteratively adding Gaussian noise to the ground truth SMPL parameters based on the given timestep. The diffusion model then learns to denoise these SMPL parameters and directly predict the ground truth parameters, as detailed in Sec.4.2.1 of the main paper.

In Fig.~\ref{fig:ablation_scheduler}, we compare the performance of models in terms of MPJPE and PVE across different data splits, using various fine-tuning scheduler strategies, including linear, cosine, exponential, and no scheduler. For the linear and cosine schedulers, the maximum number of iterations depends on the amount of real data available and the number of epochs used for fine-tuning. For the exponential scheduler, we set the multiplicative decay factor for the learning rate to $0.999$.
The results show that the linearly-adjusted scheduler achieves consistently lower errors compared to other approaches. This demonstrates the effectiveness of our fine-tuning strategy in improving the model's performance.


\subsection{Effectiveness of Synthetic Data Utilization}
\begin{figure}[ht]
  \centering
    \includegraphics[width=0.8\linewidth]{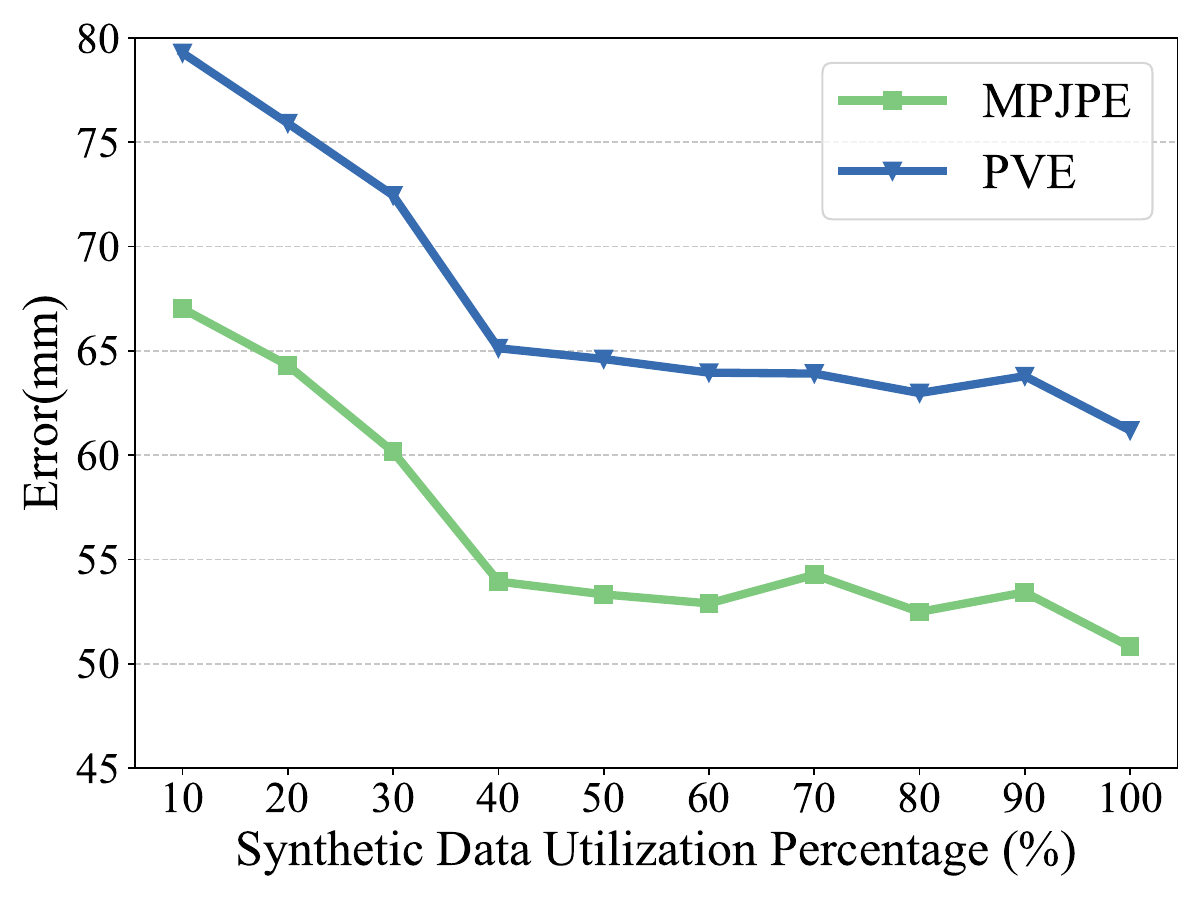}
  \caption{\textbf{Ablation Study on Synthetic Data Utilization.}}
  \label{fig:ablation_syn_ratio}
\end{figure}
In Table 1, we present experiments using all synthetic data combined with varying proportions of real training data to validate the generalizability and effectiveness of the proposed DiSRT-In-Bed pipeline. Additionally, we perform experiments to further demonstrate the impact of incorporating synthetic data. In this setting, training is conducted using all real data combined with different proportions of synthetic data, while testing is performed on the same real dataset. As shown in Fig.~\ref{fig:ablation_syn_ratio}, both MPJPE and PVE generally decrease as the proportion of synthetic data increases. However, a slight increase in error metrics is observed when synthetic data reaches 70\% and 90\% due to distribution shifts. Overall, the best performance is achieved when using all synthetic data and all real training data, as presented in Sec. 5, compared to settings with less synthetic data.

\section{Additional Visualizations}
\label{sec:sup_viz}

We present additional visualization examples to illustrate the effectiveness of our DiSRT-In-Bed method compared to the state-of-the-art BodyMAP method. As shown in Fig.~\ref{fig:viz_sup}, our proposed method achieves superior mesh predictions, especially when access to real-world data is limited. The predictions from our model align more closely with the input data and exhibit stable performance across varying covering scenarios.

Fig.~\ref{fig:viz_hospital_sup} provides additional visualizations on the SLP~\cite{slpdata} hospital setting dataset, \textbf{which features a different data distribution from the training dataset and lacks labeled ground truth}. Here, we compare our method, with and without the proposed Sim-to-Real training strategies described in Sec.4.3 of the main paper, against BodyMAP in terms of generalization to diverse real-world settings. All models were trained on the complete synthetic dataset and the full real-world SLP~\cite{slpdata} home setting dataset. 

The results reveal that our method without the Sim-to-Real training strategies performs comparably to BodyMAP; however, both are less stable across different covering scenarios and fail to capture finer details. In contrast, our proposed Sim-to-Real framework significantly enhances stability and detail alignment, demonstrating its robustness and generalization capability across varying real-world conditions.

\section{Limitations and Future Work}

\begin{figure}[ht]
  \centering
  \subcaptionbox{Self-Interpenetration.\label{fig:failure_in}}{%
    \includegraphics[width=0.9\linewidth]{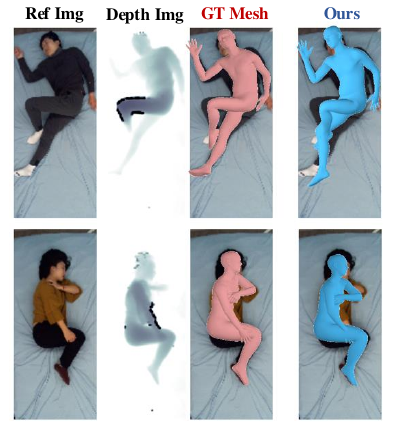}
  }
  \hspace{0.05\linewidth} 
  \subcaptionbox{Misalignment.\label{fig:failure_align}}{%
    \includegraphics[width=0.9\linewidth]{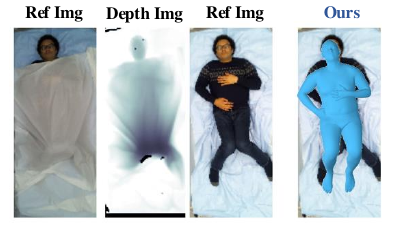}
  }
  \caption{\textbf{Failure Cases of DiSRT-In-Bed.}}
  \label{fig:failure}
\end{figure}

While our proposed DiSRT-In-Bed demonstrates promising performance in handling in-bed human mesh recovery with limited real-world data and strong generalization across different environmental settings, there are two key directions for future work: improving accuracy and enhancing scalability.

\textbf{Accuracy:} Future efforts could focus on improving the prediction quality of in-bed human body meshes. For instance, as shown in Fig.~\ref{fig:failure_in}, failure cases involving self-interpenetration remain challenging. In the first example, interpenetration occurs near the left foot and right knee due to the complex pose and the close proximity of these body parts. Similarly, in the second example, self-contact introduces ambiguity in determining the precise position of body parts. Addressing these issues could involve refining model components to better account for self-contact scenarios or incorporating additional constraints to reduce interpenetration errors.

\textbf{Scalability:} Extending DiSRT-In-Bed to establish its clinical effectiveness is another critical direction. Fig.~\ref{fig:failure_align} highlights a misaligned prediction caused by a challenging, out-of-distribution input from the SLP~\cite{slpdata} hospital-setting dataset. Addressing such misalignments in different settings could involve several approaches: expanding synthetic datasets using customizable simulations, incrementally fine-tuning the diffusion model with newly collected data, and designing new diffusion model components that integrate domain-specific knowledge. These advancements could push our framework closer to practical deployment in clinical environments.

\begin{figure*}[ht]
  \centering
    \includegraphics[width=0.9\linewidth]{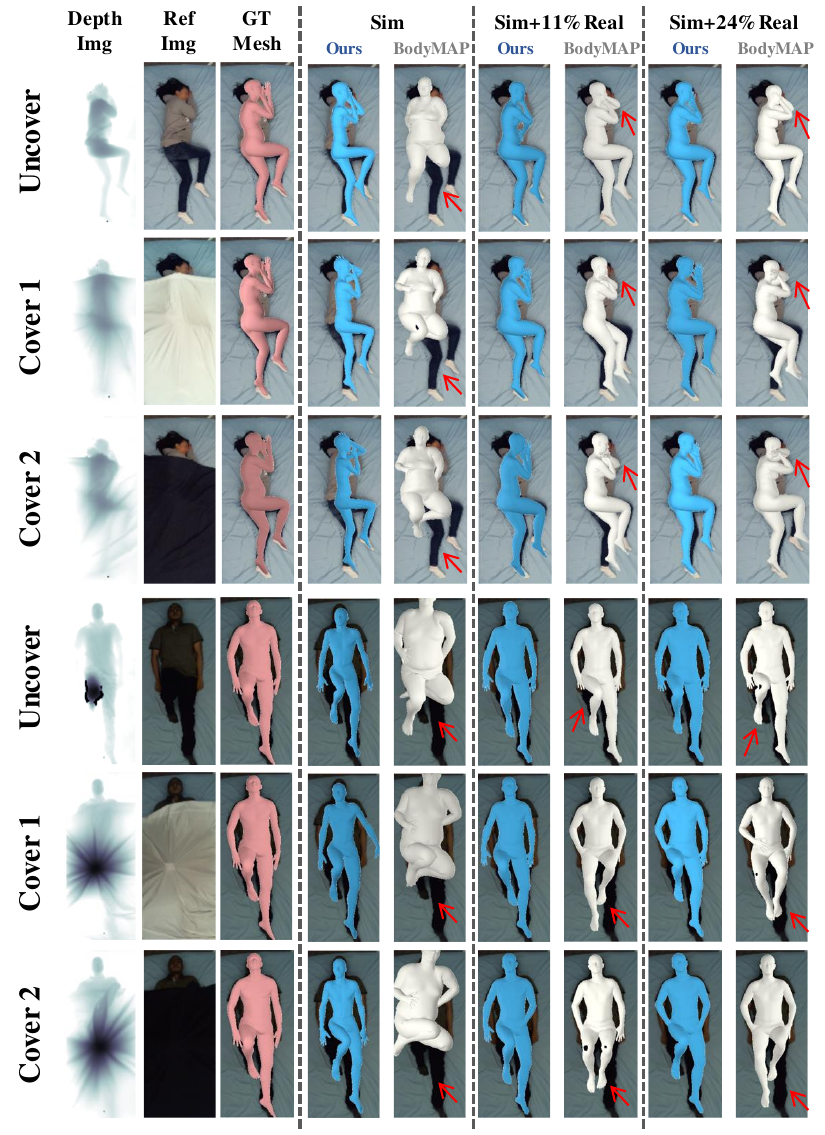}
  \caption{\textbf{Additional Visualization Comparison with Baseline on the SLP~\cite{slpdata} Home-Setting Dataset.}}
  \label{fig:viz_sup}
\end{figure*}

\begin{figure*}[ht]
  \centering
    \includegraphics[width=0.9\linewidth]{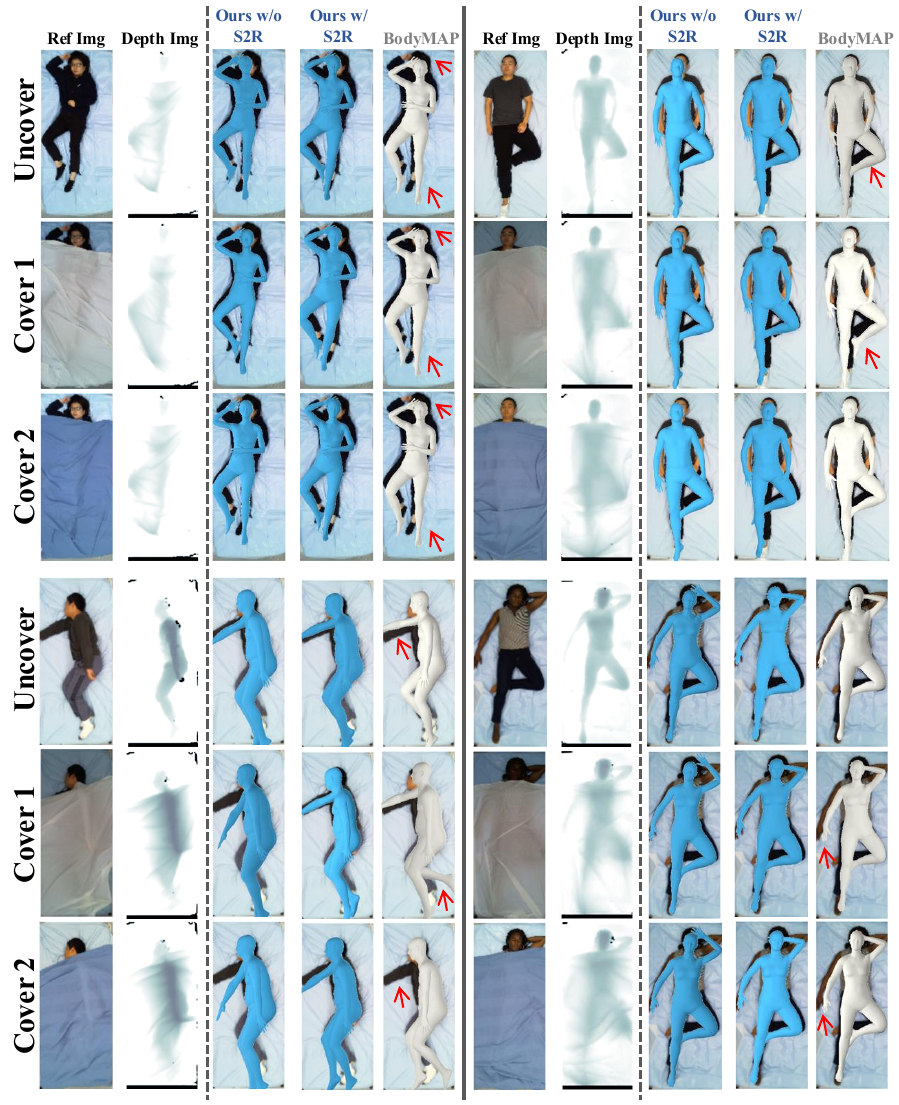}
  \caption{\textbf{Additional Visualization Comparison with Baseline on the SLP~\cite{slpdata} Hospital-Setting Dataset.}}
  \label{fig:viz_hospital_sup}
  \vspace{5pt}
\end{figure*}

\end{document}